\PassOptionsToPackage{dvipsnames,table}{xcolor}

\documentclass[10pt,twocolumn,letterpaper]{article}

\usepackage[pagenumbers]{cvpr} % To force page numbers, e.g. for an arXiv version

\definecolor{cvprblue}{rgb}{0.21,0.49,0.74}
\usepackage[pagebackref,breaklinks,colorlinks,allcolors=cvprblue]{hyperref}
\usepackage{algorithm}
\usepackage[abbreviations]{foreign}
\usepackage{booktabs}
\usepackage{multirow}
\usepackage{subcaption}
\usepackage{amsmath}   % 提供了 \boldsymbol 等
\usepackage{amsfonts}  % 提供了 \mathbb
\usepackage{amssymb}   % 也会加载 amsfonts，并提供更多符号
\usepackage{mathrsfs}   % 提供 \mathscr
\usepackage{amsthm}     

\usepackage{nicefrac}
\usepackage{algpseudocode}
\DeclareMathOperator{\Var}{\mathrm{Var}}

\usepackage{newfloat}
\usepackage{listings}
\def\confName{CVPR}
\def\confYear{2025}

\title{Quality-Aware Language‑Conditioned Local Auto‑Regressive\\Anomaly Synthesis and Detection}

\author{Long Qian$^{1,2}$~~~~
Bingke Zhu$^{1,2}$~~~~
Yingying Chen$^{1,2}$~~~~
Ming Tang$^{1,2}$~~~~
Jinqiao Wang$^{1,2,3}$\\
  $^{1}$~Foundation Model Research Center, Institute of Automation, \\ Chinese Academy of Sciences, Beijing, China \\ 
  $^{2}$~School of Artificial Intelligence, University of Chinese Academy of Sciences, Beijing, China\\
  $^{3}$~Objecteye Inc., Beijing, China\\
  {\tt\small  qianlong2024@ia.ac.cn} \\
  {\tt\small \{bingke.zhu,yingying.chen,tangm,jqwang\}@nlpr.ia.ac.cn}
}

\begin{document}
\maketitle
\begin{abstract}
Despite substantial progress in anomaly synthesis methods, existing diffusion-based and coarse inpainting pipelines commonly suffer from structural deficiencies such as micro-structural discontinuities, limited semantic controllability, and inefficient generation. To overcome these limitations, we introduce \textit{ARAS}, a language-conditioned, auto-regressive anomaly synthesis approach that precisely injects local, text-specified defects into normal images via token-anchored latent editing. Leveraging a hard-gated auto-regressive operator and a training-free, context-preserving masked sampling kernel, \textit{ARAS} significantly enhances defect realism, preserves fine-grained material textures, and provides continuous semantic control over synthesized anomalies. Integrated within our Quality-Aware Re-weighted Anomaly Detection (\textit{QARAD}) framework, we further propose a dynamic weighting strategy that emphasizes high-quality synthetic samples by computing an image-text similarity score with a dual-encoder model. Extensive experiments across three benchmark datasets—\textit{MVTec~AD}, \textit{VisA}, and \textit{BTAD}, demonstrate that our \textit{QARAD} outperforms \textbf{SOTA} methods in both image- and pixel-level anomaly detection tasks, achieving improved accuracy, robustness, and a 5× synthesis speedup compared to diffusion-based alternatives. Our complete code and synthesized dataset will be publicly available.
\end{abstract}    
\section{Introduction}
\label{sec:intro}

Anomaly detection remains critically challenged by the persistent and fundamental issue of \emph{data imbalance}: while massive amounts of normal data are collected, anomalous samples are rare, and difficult to gather comprehensively in real-world environments. To bridge this gap, recent efforts have increasingly turned toward anomaly synthesis, utilizing powerful generative frameworks such as \textit{Augmentation-based}~\cite{li2021cutpaste, liu2023simplenetsimplenetworkimage, Schluter2022NSA}, \textit{GANs}~\cite{Perera2019OCGAN, Akcay2018GANomaly}, and diffusion-based models~\cite{Wyatt2022AnoDDPM, he2023diaddiffusionbasedframeworkmulticlass}. Despite significant progress, recent synthesis methods suffer from several pervasive limitations.

\begin{figure*}[t]
    \centering
    \includegraphics[width=\linewidth]{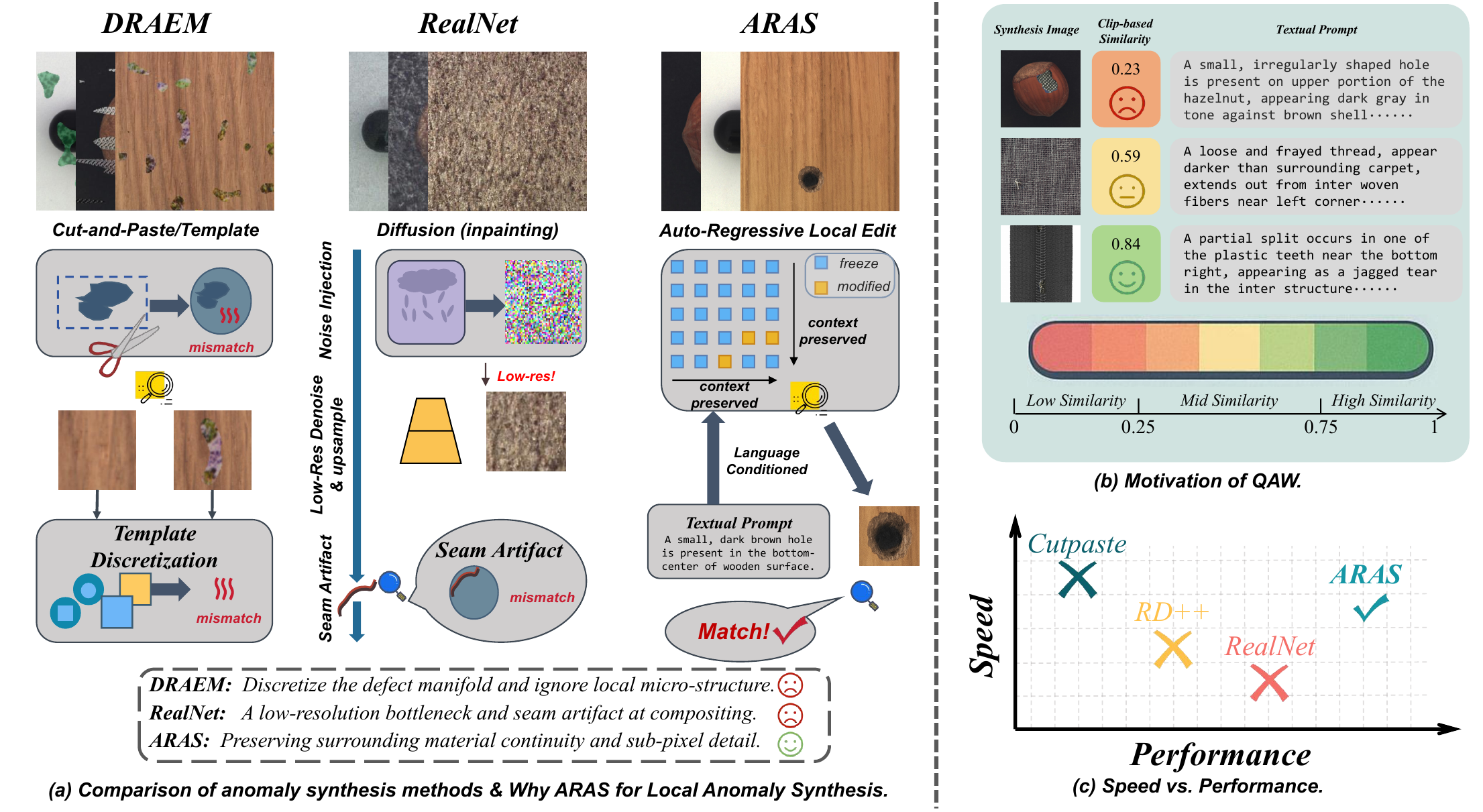}
    \caption{\textbf{Motivation of \textit{ARAS} and Quality-Aware Weighting (\textit{QAW})} 
      \textbf{(a) Coarse anomaly synthesis vs.\ \textit{ARAS}.}
      \textit{Cut-Paste}/\textit{Template} and diffusion inpainting pipelines either discretize the defect manifold and ignore local micro-structure or pass through a noisy, low-resolution denoising bottleneck that yields seam artifacts when composited. Our mask-conditioned, language-guided auto-regressive local edit (\textit{ARAS}) freezes surrounding context tokens and inserts fine-grained defects aligned with material texture.
      \textbf{(b) Sample quality varies.} Text-image alignment (\textit{CLIP} similarity) reveals that many synthetic samples fail to express their prompts.
      \textbf{(c) Speed vs.\ performance.} \textit{ARAS} achieves substantially faster synthesis than diffusion-based methods while improving downstream anomaly detection accuracy.
    }
    \label{fig:motivation}
\end{figure*}

Recent diffusion-based anomaly synthesis methods~\cite{Zhang2024RealNet,Wyatt2022AnoDDPM} typically operate \emph{globally} or via \emph{coarse-grained inpainting}. A representative pipeline encodes the entire image (or masked crop) into a noisy latent, progressively denoises at a reduced working resolution, and finally upsamples and composites the region back into the original image. Traditional methods~\cite{10.1007/978-3-031-19821-2_31, Zavrtanik2021DRAEM,Tien_2023_CVPR} similarly rely on low-frequency residual modulation or random blob. While effective for gross structural corruption, these schemes struggle to \emph{precisely respect the micro-structure of the underlying material}. See Fig.~\ref{fig:motivation}(a) which illustrates several recurring \emph{symptoms} (\eg texture breaks and seam artifact), but the deeper \emph{causal mechanisms} matter for designing better generators.

\emph{Recent methods suffer from high‐noise overwriting and a resolution bottleneck.} Coarse inpainting overwrites masked pixels with high-variance noise latents, denoises at a spatially compressed resolution, then upsamples. Each stage discards sub-pixel structure (\eg metal grain, fiber weave). When the patch is fused back, phase mis-match emerges at the boundary. This is a \emph{structural} failure: fine-scale material statistics never propagate through the noisy, low-res bottleneck. As seen in Fig.~\ref{fig:motivation}(a), diffusion-based and traditional methods exhibit such artifacts. \emph{Template discretization restricts the semantic granularity of synthesized defects.} Many pipelines provide only categorical control (\eg ``scratch'', ``stain'', or random blob masks~\cite{li2024promptadlearningpromptsnormal}). This discretizes a \emph{continuous} defect manifold (\eg color shift, alignment to texture flow) into a handful of coarse prototypes \emph{uninformed by the actual local micro-structure}. As a result, synthesized anomalies rarely conform to material-dependent cues that real detectors must learn. \emph{Uniform training weights amplify the influence of poor‑quality synthetic data.} Current detectors typically treat all synthetic samples equally during optimization. However, generations that deviate from their intended semantics (ambiguous prompt, collapsed texture) are often easier to fit and can produce disproportionately large gradients, steering the model toward artifactual cues. This \emph{``bad drives out good''} effect degrades detection reliability. Fig.~\ref{fig:motivation}(b) visualizes our motivation of \textit{QAW}.

Motivated by these critical gaps, we introduce a language‑conditioned auto‑regressive patch editor that synthesize anomaly region: \emph{\textbf{A}uto-\textbf{R}egressive \textbf{A}nomaly \textbf{S}ynthesis} (\textit{\textbf{ARAS}}), seamlessly integrated into our broader \emph{\textbf{Q}uality‑\textbf{A}ware \textbf{R}e‑weighted \textbf{A}nomaly \textbf{D}etection} (\textit{\textbf{QARAD}}) framework. At its core, \textit{ARAS} leverages the \textit{Infinity} architecture~\cite{Infinity}, a powerful \textit{8B AR} model operating on vector-quantized variational autoencoder (\textit{VQ-VAE}) latent tokens. In contrast to prior coarse methods, \textit{ARAS} precisely locks all tokens outside user-specified anomaly masks, ensuring only desired local pixels are synthesized. Coupled with detailed linguistic prompts, \textit{ARAS} enables semantic and spatial control, synthesizing defects described by textual descriptions, while faithfully preserving the intricate textures and micro-structural fidelity of original images. Unlike previous approaches that implicitly learn textures through iterative refinement, \textit{ARAS} explicitly leverages context-aware latent anchoring and linguistic prompts, effectively encoding material-specific priors into the anomaly synthesis process.

Embedded within our \textit{QARAD} framework, we propose a novel \emph{Quality-Aware Weighting} (\textit{QAW}) strategy.  Rather than uniformly treating all synthetic anomalies, \textit{QAW} dynamically adjust the training weight of each synthetic sample based on the semantic consistency between its linguistic description and generated visual representation. Specifically, leveraging \textit{CLIP}-based similarity scores, \textit{QAW} systematically down-weight low-quality synthetic samples, mitigating their negative impact on the anomaly detector’s training which substantially enhances detection stability and accuracy. This continuous re‑weighting couples synthesis and detection in a single loop, improving robustness without discarding data diversity. Moreover, by adaptively calibrating the gradient landscape during training, \textit{QAW} encourages the detector to prioritize subtle, high-fidelity cues over misleading coarse artifacts, yielding detectors that generalize significantly better to real-world anomalies.

Across \textit{MVTec~AD}, \textit{VisA}, and \textit{BTAD} benchmarks, \textit{QARAD} consistently improves image- and pixel-level AUROC over diffusion-style synthesis baselines and traditional methods. Because we avoid iterative denoising, \textit{ARAS} synthesizes a $1024^2$-resolution defect image in a single forward pass, reducing per-sample generation latency by at least 5 times relative to diffusion inpainting (see Fig.~\ref{fig:motivation}(c)).

In summary, our contributions are three-fold:

\begin{itemize}
    \item We propose \emph{ARAS}, the first auto-regressive, language-driven anomaly synthesis method capable of precise, pixel-level local editing of defects in high-resolution industrial images, significantly surpassing traditional approaches in realism, controllability, and efficiency.
    
    \item We introduce the comprehensive anomaly detection framework \emph{QARAD}, which integrates \textit{ARAS} with a novel Quality-Aware Weighting scheme to selectively emphasize high-quality synthetic samples, thus stabilizing training and improving detection performance.
    
    \item Evaluations across datasets, \textit{MVTec~AD}, \textit{VisA} and \textit{BTAD}, demonstrate that \textit{QARAD} achieves \textbf{SOTA} performances.
\end{itemize}

\section{Related Work}

\paragraph{Synthesis-Based Industrial Anomaly Detection.}
Because defective samples are scarce in manufacturing, many \textit{AD} systems learn from synthetic anomalies composited onto normal imagery. Early but still influential industrial baselines such as \textit{CutPaste}~\cite{li2021cutpaste} paste randomly cropped texture patches to simulate defects, improving data diversity for one‑class detectors. \textit{DRAEM}~\cite{Zavrtanik2021DRAEM} extends this idea with dual reconstruction and segmentation branches trained on synthetic corruptions blended into normal images, establishing a strong pixel‑level \textit{AD} benchmark. Recent work has pushed toward more realistic or targeted corruptions: \textit{RD++}~\cite{Tien_2023_CVPR} augments residual defects at multiple scales to sharpen boundary cues and improve localisation. Diffusion models have recently been adapted for industrial anomaly synthesis; \textit{RealNet}~\cite{Zhang2024RealNet} leverages a diffusion‑driven synthetic defect generator paired with a feature selection network to better match real failure texture statistics. \textit{AnomalyDiffusion}~\cite{Hu2024AnomalyDiffusion} further conditions a denoising diffusion process on structural priors to inject class‑aware yet diverse anomalous patterns for industrial inspection. Other approaches like \textit{SPADE}~\cite{yoon2022spadesemisupervisedanomalydetection},  \textit{DiffusionAD}~\cite{zhang2023diffusionad}, \textit{TransFusion}~\cite{Fucka2024TransFusion}, \textit{GLAD}~\cite{yao2024gladbetterreconstructionglobal} also demonstrate the utility of synthetic data, yet most operate through stochastic global corruption or coarse inpainting of large regions, which can blur fine material microstructure that is critical for anomalies, motivating our focus on mask‑local, high‑fidelity synthesis.

\paragraph{Language-Conditioned Anomaly Synthesis and Detection.}
Natural‑language supervision offers a scalable channel for specifying rare or long‑tail defect semantics beyond fixed class taxonomies. \textit{PromptAD}~\cite{li2024promptadlearningpromptsnormal} explores using textual prompts to guide anomaly generation and improve category transferability in industrial \textit{AD} settings. More broadly, open‑vocabulary and \textit{multimodal AD} systems like \textit{AnomalyGPT}~\cite{gu2023anomalygptdetectingindustrialanomalies} are emerging: \textit{AnomalyAny}~\cite{sun2025unseenvisualanomalygeneration} leverages large text–image models to support free‑form prompt queries across multiple industrial datasets and shows that textual conditioning can retrieve or localise defect evidence without per‑class retraining. Despite this progress, current language‑aware pipelines (\eg \textit{AdaCLIP}~\cite{Cao_2024}, \textit{AnomalyCLIP}~\cite{zhou2025anomalyclipobjectagnosticpromptlearning}, \textit{AACLIP}~\cite{ma2025aaclipenhancingzeroshotanomaly} and \textit{VCPCLIP}~\cite{qu2024vcpclipvisualcontextprompting}) either map prompts to coarse class tokens or apply text conditioning at image‑global scale; none tightly couple fine‑grained spatial masks with rich textual attributes (size, color shift), which our \textit{ARAS} framework addresses.

\paragraph{Auto-regressive Visual Generative Models.}
Large‑scale auto-regressive (\textit{AR}) decoders have recently re‑emerged as competitive, scalable image generators. \textit{VAR}~\cite{VAR} shows that token‑factorized visual \textit{AR} models achieve strong fidelity and flexible conditioning, and further explores visual auto-regressive modelling for detection tasks.  \textit{Infinity}~\cite{Infinity} scales this paradigm to multi‑billion‑parameter transformers operating directly over vector‑quantized visual tokens, supports dynamic resolution schedules, and exposes interfaces for selective token resampling—properties that make it naturally suited to local masked editing at test time. Our \textit{ARAS} synthesis engine builds on these \textit{AR} advances: by freezing all non‑masked VQ tokens and sampling only the masked subset under language guidance, we produce high‑detail, spatially targeted industrial defects substantially faster than iterative diffusion pipelines while preserving surrounding context.

\section{Method}
\label{sec:method}

\begin{figure*}[t]
    \centering
    \includegraphics[width=\linewidth]{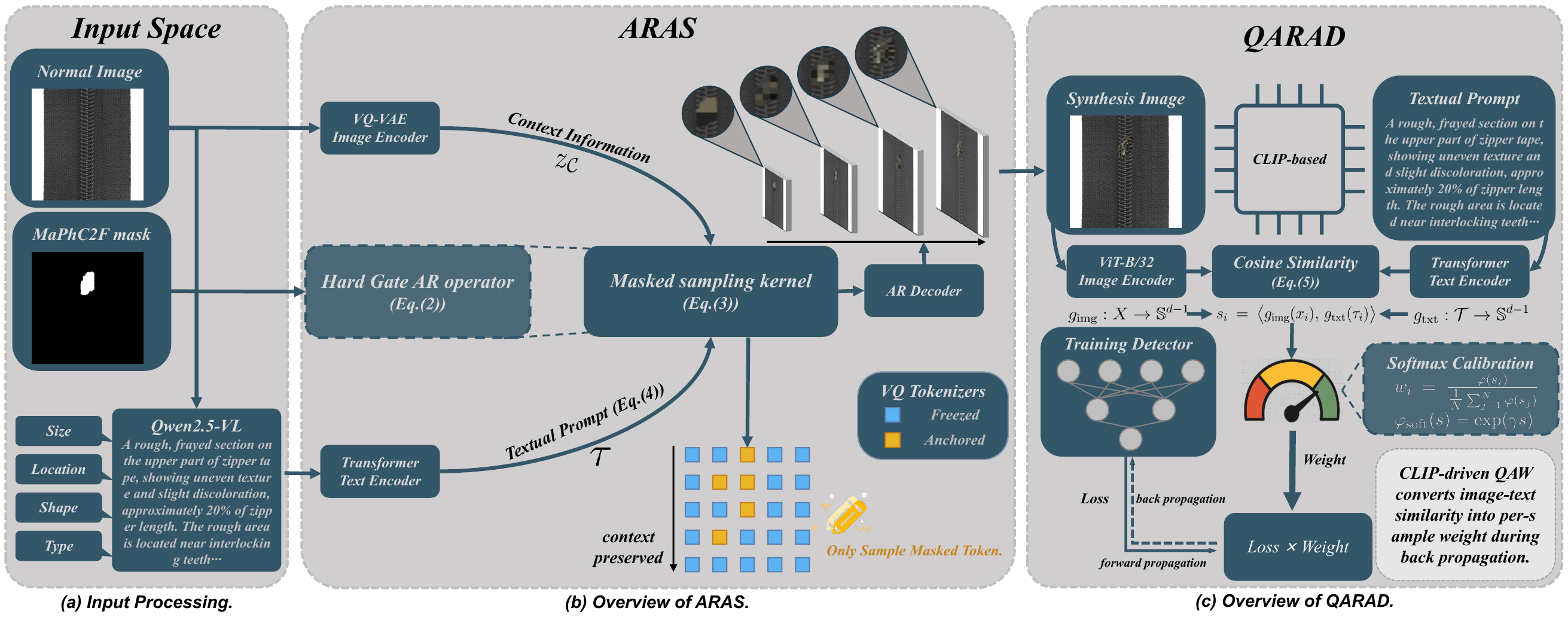}
    \caption{\textbf{End-to-End Pipeline of \textit{ARAS} and \textit{QARAD} Framework}.
    (a)~\emph{Input Processing}: raw image–mask-text triplets are generated and act as the subsequent input of our \textit{ARAS} and \textit{QARAD}; 
    (b)~\emph{ARAS} inserts linguistically‑specified local defects by freezing all context \textit{VQ‑tokens} and sampling only those indexed by the anomaly mask.  
    (c)~\emph{QARAD} down‑weights prompt‑inconsistent synthesis images by scaling each sample’s loss with a \textit{CLIP}‑based image–text similarity score.}
    \label{fig:arad_method}
\end{figure*}

\subsection{Overview}
\label{ssec:overview}

Fig.~\ref{fig:arad_method} depicts our \emph{two‑stage yet fully decoupled} pipeline.  
Let  

\[
x\in\mathbb{R}^{H\times W\times3},\quad
m\in\{0,1\}^{H\times W},\quad
\tau\in\mathcal{T},
\]

\noindent denote an original image, a binary anomaly mask, and a fine‑grained text description, respectively.  
We introduce two deterministic operators:

\[
\underbrace{\mathcal{G}_{\theta^\star}\!:\;(x,m,\tau)\;\longmapsto\;\widehat{x}}
_{\text{\textit{ARAS}}}
\quad\text{and}\quad
\underbrace{\mathcal{W}\!:\;(\widehat{x},\tau)\;\longmapsto\;w\in\mathbb{R}_{\!+}}
_{\text{\textit{QARAD}}},
\]

\noindent where $\mathcal{G}_{\theta^\star}$ is a frozen large‑scale auto‑regressive editor that \emph{replaces} the token subset indexed by $m$ while \emph{identity‑mapping} the complement; $\mathcal{W}$ assigns a reliability weight to each synthesis via an image–text alignment score.

Denoting by $\mathcal{D}_{\mathrm{r}}$ the distribution of real samples and by $\mathcal{D}_{\mathrm{s}}$ the sampling measure over $(x,m,\tau)$ triplets, called $\mathcal{A}$, the detector $f_\phi$ is obtained by minimising the \emph{weighted risk}

\begin{align}
\label{eq:global_obj}
\min_{\phi}
&\mathbb{E}_{\mathcal{A}\sim\mathcal{D}_{\mathrm{s}}}
    \Bigl[
      \underbrace{\mathcal{W}\bigl(\mathcal{G}_{\theta^\star}\mathcal{A},\;\tau\bigr)}_{w}
      \;\cdot\;
      \ell\bigl(f_\phi(\mathcal{G}_{\theta^\star}\mathcal{A}),\;1\bigr)
    \Bigr]  \nonumber\\
&+\mathbb{E}_{(x,y)\sim\mathcal{D}_{\mathrm{r}}}
    \bigl[\ell(f_\phi(x),\,y)\bigr]
\end{align}

\noindent with loss $\ell$. Eq.~\eqref{eq:global_obj} formalises the \textbf{decoupled design Philosophy}: \emph{Generator modularity}, $\theta^\star$ is frozen; any stronger \textit{AR} editor may replace $\mathcal{G}$ without touching detector training. \emph{Quality adaptivity}, $\mathcal{W}$ down‑weights prompt–inconsistent syntheses while preserving their diversity.

\subsection{\textit{ARAS}: Token‑Anchored, Training‑Free Local Edit}
\label{ssec:aras}

\paragraph{Token lattice.}
A normal image $x$ is discretised by a fixed vector‑quantiser
$E:\mathbb{R}^{H\times W\times3}\!\!\rightarrow\!\mathbb{Z}^{h\times w}$,
yielding a latent lattice
$
z = E(x) = \bigl\{z_{ij}\bigr\}_{i,j}.
$
Let the binary anomaly mask be $m\!\in\!\{0,1\}^{H\times W}$ and define the token index sets
\[
\mathcal{C} \;=\;\bigl\{(i,j)\mid m_{ij}=0\bigr\}, \quad
\mathcal{M} \;=\;\bigl\{(i,j)\mid m_{ij}=1\bigr\}.
\]

\paragraph{Hard‑gated auto-regressive operator.}
Denote by $p_{\vartheta}$ a \emph{frozen} auto-regressive decoder over the token space.
We introduce a \emph{hard‑gating} functional  
\begin{equation}
\label{eq:gating}
\mathsf{Gate}_{\mathcal{C}}(q)
\;:=\;
\prod_{(i,j)\in\mathcal{C}}
    \delta\bigl[q(z_{ij}) - z_{ij}\bigr],
\end{equation}
where $\delta$ is the \textit{Dirac mass} and $q$ an arbitrary joint distribution on $z$.  
Eq.~\eqref{eq:gating} forces unit probability on context tokens.

\paragraph{Masked‑AR kernel.}
Let $\pi$ be a bijection $\{1,\dots,|\mathcal{M}|\}\!\to\!\mathcal{M}$.
We define the \emph{token‑anchored sampling kernel}
\begin{equation}
\label{eq:masked_kernel}
    \mathcal{K}_{\tau,\mathcal{C}} \;=\; \mathsf{Gate}_{\mathcal{C}} \Bigl(\prod_{t=1}^{|\mathcal{M}|}p_{\vartheta}\bigl(z_{\pi(t)} \;\bigl|\;z_{\mathcal{C}},z_{\pi(<t)},\tau\bigr)\Bigr)
\end{equation}
which leaves $z_{\mathcal{C}}$ untouched and successively samples masked positions under prompt~$\tau$, context $z_{\mathcal{C}}$ and tokens $z_{\pi(<t)}$.
Sampling produces $\widehat z\!=\!\mathcal{K}_{\tau,\mathcal{C}}(z)$, subsequently decoded by $D$ to obtain the locally edited image $\widehat x=D(\widehat z)$.

\textbf{Context‑invariance guarantee.}
Immediate from the definition of $\mathsf{Gate}_{\mathcal{C}}(\cdot)$ in Eq.~\eqref{eq:gating}, for any $(x,m,\tau)$ and for any stochastic realisation of $\widehat z\!\sim\!\mathcal{K}_{\tau,\mathcal{C}}(z)$,
\(
\widehat z_{ij}=z_{ij}\;\,\forall\,(i,j)\!\in\!\mathcal{C}.
\)
Hence $\widehat x_{\,\mathcal{C}}=x_{\,\mathcal{C}}$, we guarantee context‑invariance.

\paragraph{Prompt‑conditioned micro‑structural field.}
Let the text prompt be encoded as a point 
$\boldsymbol{\tau}\!\in\!\mathcal{T}\subset\mathbb{R}^{d}$,
a smooth \emph{semantic manifold}.
Eq.~\eqref{eq:masked_kernel} induces a stochastic operator
\[
\Phi:\;
\mathcal{T}\times\mathbb{Z}^{h\times w}
\;\longrightarrow\;
\mathcal{P}\!\bigl(\mathbb{Z}^{|\mathcal{M}|}\bigr),\qquad
(\,\boldsymbol{\tau},\,z_{\mathcal C})\;\mapsto\;
p_{\boldsymbol{\tau},z_{\mathcal C}},
\]
where $\mathcal{P}(\cdot)$ denotes the probability simplex.  
Under the cross‑modal key–value projection of the \textit{AR} decoder,  
$\Phi$ is \emph{Lipschitz‑continuous} in $\boldsymbol{\tau}$;  
an infinitesimal prompt edit
(\eg defect length $+\varepsilon$\,mm)  
perturbs the token logits by at most $L\varepsilon$ for some
$L\!<\!\infty$.  
Sampling
\begin{equation}
\label{eq:text_field}
    \widehat z_{\mathcal{M}}
    \;\sim\;
    p_{\boldsymbol{\tau},z_{\mathcal C}}
    \qquad\Longrightarrow\qquad
    \widehat z_{\mathcal{M}}
    =\mathscr{F}_{\tau}(z_{\mathcal C})
\end{equation}

\noindent therefore delivers a \textbf{semantically differentiable} field
that \emph{inherits} the stationary high‑frequency statistics
(grain, weave, gloss) of the frozen context $z_{\mathcal C}$.

In contrast, any method that re‑encodes the entire patch (downsample $\!\rightarrow$ denoise $\!\rightarrow$ upsample) breaks this coupling and forfeits micro‑structural coherence.

Our \textit{ARAS} converts a generic auto-regressive decoder into a \emph{local, prompt‑selective anomaly synthesis editor} via the hard‑gating mechanism in Eq.~\eqref{eq:gating} and the masked‑kernel operator of Eq.~\eqref{eq:masked_kernel}. The result is a \textit{\textbf{training‑free}, \textbf{micro‑structure preserving}} anomaly injector that enjoys exact locality and continuous attribute control through~$\tau$. These properties jointly address the structural deficiencies of augmentation‑based methods (semantic‑granularity loss) and diffusion‑based methods (phase discontinuity).

\subsection{\textit{QARAD}: Quality‑Aware Re‑Weighting}
\label{ssec:qarad}

Synthetic anomalies, while diverse, vary in their fidelity to the user‑specified semantics.  To prevent low‑consistency outliers from dominating training, \textit{QARAD} endows each synthetic sample with a continuous \emph{reliability weight} \(w_i\) based on its image–text alignment, and integrates these weights into an importance‑weighted risk estimator.

\paragraph{Consistency scoring as pseudo‑density.}
Define a dual‑encoder mapping into the unit hypersphere,
\[
g_{\text{img}}:X\to\mathbb{S}^{d-1},\quad
g_{\text{txt}}:\mathcal{T}\to\mathbb{S}^{d-1}.
\]
For each pair \((x_i,\tau_i)\), compute the cosine similarity
\begin{equation}
\label{eq:sim_c}
s_i \;=\;\bigl\langle g_{\text{img}}(x_i),\,g_{\text{txt}}(\tau_i)\bigr\rangle\;\in[-1,1].
\end{equation}
We interpret \(s_i\) as an unnormalised proxy for the \emph{likelihood ratio} between the true conditional defect distribution and the sampler’s output, thereby guiding the weighting mechanism to prioritise semantically faithful samples.

\begin{table*}[t]
\centering
\begin{tabular}{c|cccccc}
\toprule
\textit{Category} & \textit{DSR}& \textit{SimpleNet}& \textit{DRAEM}&\textit{RD++}& \textit{RealNet}& \textit{QARAD~(Ours)}
\\
\midrule
\textit{Candle}& 86.4/79.7& 92.3/97.7& 91.8/96.6&96.4/98.6& 96.1/99.1& \textbf{97.8}/\textbf{99.9}\\
\textit{Capsules} & 93.4/74.5& 76.2/94.6& 74.7/98.5&92.1/99.4& 93.2/98.7&\textbf{97.4}/\textbf{99.8}\\
\textit{Cashew} & 85.2/61.5& 94.1/99.4& 95.1/83.5&97.8/95.8& 97.8/98.3&\textbf{98.6}/\textbf{99.9}\\
\textit{Chewinggum} & 97.2/58.2& 97.1/97.0& 94.8/96.8&96.4/99.0& \textbf{99.9}/99.8&\textbf{99.9}/\textbf{100}\\
\textit{Fryum} & 93.0/65.5& 88.0/93.5& 97.4/87.2&95.8/94.3& 97.1/96.2&\textbf{99.4}/\textbf{99.9}\\
\textit{Macaroni1} & 91.7/57.7& 84.7/95.4& 97.2/\textbf{99.9}&94.0/99.7& \textbf{99.8}/\textbf{99.9}&99.4/\textbf{99.9}\\
\textit{Macaroni2} & 79.0/52.2& 75.0/83.8& 85.0/99.2&88.0/87.7& 95.2/99.6&\textbf{97.4}/\textbf{99.8}\\

\textit{PCB1} & 89.1/61.3& 93.4/99.1& 47.6/88.7&97.0/75.0& \textbf{98.5}/\textbf{99.7}&\textbf{98.5}/99.4\\
\textit{PCB2} & 96.4/84.9& 90.0/94.8& 89.8/91.3&97.2/64.8& 97.6/98.0&\textbf{99.4}/\textbf{99.9}\\
\textit{PCB3} & 97.0/79.5& 91.3/98.2& 92.0/98.0&96.8/95.5& \textbf{99.1}/98.8&\textbf{99.1}/\textbf{99.9}\\
\textit{PCB4} & 98.5/62.1& 99.1/94.5& 98.6/96.8&\textbf{99.8}/92
.8& 99.7/98.6&99.7/\textbf{99.7}\\
\textit{Pipe fryum} & 94.3/80.5& 89.0/95.3& \textbf{100}/85.8&99.6/92.0& 99.9/99.2&99.9/\textbf{99.9}\\
\midrule

\rowcolor[gray]{0.9}
\textit{\textbf{Avg.}}    & 91.8/68.1& 89.2/95.3& 88.7/93.5&95.9/90.1& 97.8/98.8&\textbf{98.9}/\textbf{99.8}\\
\bottomrule
\end{tabular}

\caption{Performance comparison across different SOTA methods on \textit{VisA} dataset. \textbf{Bold text} indicates the best performance among all method. The values in the form of \textit{xx/xx} represent \textit{image-level AUROC / pixel-level AUROC}.}
\label{tab:visa}
\end{table*}

\paragraph{Monotonic calibration \(\varphi\).}
We then pass \(\{s_i\}\) through a smooth, strictly increasing calibration function
\[
\varphi:\;[-1,1]\;\longrightarrow\;\mathbb{R}_{+},
\]
to obtain weights
\begin{equation}
\label{eq:weight_general}
w_i \;=\;\frac{\varphi(s_i)}{\tfrac1N\sum_{j=1}^N\varphi(s_j)},
\qquad
\mathbb{E}[w_i]=1.
\end{equation}
Two effective choices are:
\[
\varphi_{\mathrm{soft}}(s)=\exp(\gamma s),
\quad
\varphi_{\mathrm{hinge}}(s)=\max\{0,\,s-\beta\},
\]
each offering a tunable parameter (\(\gamma\) or \(\beta\)) that sharpens the weight distribution around high‑quality samples. We empirically adopt the \textit{softmax} form because it down-weights low-consistency samples \emph{smoothly} while granting them non-zero influence, thus retaining sample diversity and stabilising early optimization, whereas \textit{hinge} can discard too much gradient signal when the quality score distribution is narrow.

\paragraph{Variance reduction guarantee.} We will prove it as follows.
Let $\ell_i\!=\!\ell\!\bigl(f_\phi(\mathcal{G}(x_i,m_i,\tau_i)),1\bigr)$ 
and $w_i$ be defined as in Eq.~\eqref{eq:weight_general}.
If the loss $\ell_i$ is \emph{positively correlated} with any monotone
function of the similarity score $s_i$ (\ie\ $\mathrm{Cov}\!\bigl(\ell_i,f(s_i)\bigr)\!>\!0$),
then classical importance‑sampling theory ~\cite{mcbook} immediately yields
\[
\Var\!\bigl[w_i\,\ell_i\bigr] \;\le\; \Var[\ell_i],
\]
with strict inequality whenever $\ell_i$ and $s_i$ are not independent.
Hence, weighting by $\varphi(s)$ \emph{always reduces—or in the worst case, leaves unchanged—the variance of the synthetic risk term}, providing a formal justification for the empirical stability we observe, and guarantee variance reduction.

The proposed weighting scheme is \emph{unbiased}, because the normalization $\mathbb{E}[w_i]=1$ preserves the expected risk; \emph{adaptive}, since high‑consistency samples are automatically assigned larger gradient amplitudes and low‑consistency ones are softly down‑scaled; and \emph{decoder‑agnostic}, as any image–text dual encoder can replace $(g_{\mathrm{img}},g_{\mathrm{txt}})$ without altering the formulation, \emph{thereby providing a lightweight drop‑in upgrade with virtually few computational overhead}.

Together, \textit{ARAS} and \textit{QARAD} form a fully \emph{training‑free generator + quality‑adaptive learner} pipeline: \textit{ARAS} supplies high‑fidelity, micro‑structure‑aware anomalies, while \textit{QARAD} ensures that only those samples which faithfully realize their textual intent dominate detector optimization.

\section{Experiments}
\label{sec:exp}
\subsection{Experimental Setup}
\label{ssec:exp_setup}

\paragraph{Datasets.}
We conduct all experiments on three public industrial anomaly detection benchmarks:  
\emph{MVTec~AD} ($15$ categories, $\sim$$5.3$ k images)~\cite{bergmann2019mvtec}, \emph{VisA} ($12$ categories, $\sim$$10.8$ k images)~\cite{zou2022spot}, and \emph{BTAD} ($3$ categories, $\sim$$2.5$ k images)~\cite{9576231}.

\paragraph{Evaluation metrics.}
Performance is reported with the two community standards: \emph{Image‑level AUROC} and \emph{Pixel‑level AUROC}. Higher is better for both evaluation metrics.

\paragraph{Implementation details.}
For each normal training image we sample \textit{6} anomaly masks from the \emph{MaPhC2F} dataset~\cite{qian2025mathphysguidedcoarsetofineanomalysynthesis} and generate a fine‑grained prompt using \textsc{Qwen2.5‑VL‑32B‑Instruct}~\cite{Qwen2.5-VL},  
which is then passed to our \emph{training‑free} \textit{ARAS} editor, developed by the public \textsc{Infinity‑8B} auto‑regressive architecture~\cite{Infinity}; all context tokens are hard‑gated and only the masked \textit{tokens} are resampled by our \textit{masked-AR Kernel}, yielding a locally edited anomaly image.

For anomaly detection we adopt \textit{RealNet} as the backbone and inject our \textit{QAW} module: \textit{CLIP} encodes the anomaly image and textual prompts to compute the similarity score; weights are obtained via the \textit{softmax} calibration.

\subsection{Comparison with SOTAs}

\begin{figure}[h]
    \centering
    \begin{subfigure}[b]{0.95\linewidth}
        \centering
        \includegraphics[width=\linewidth]{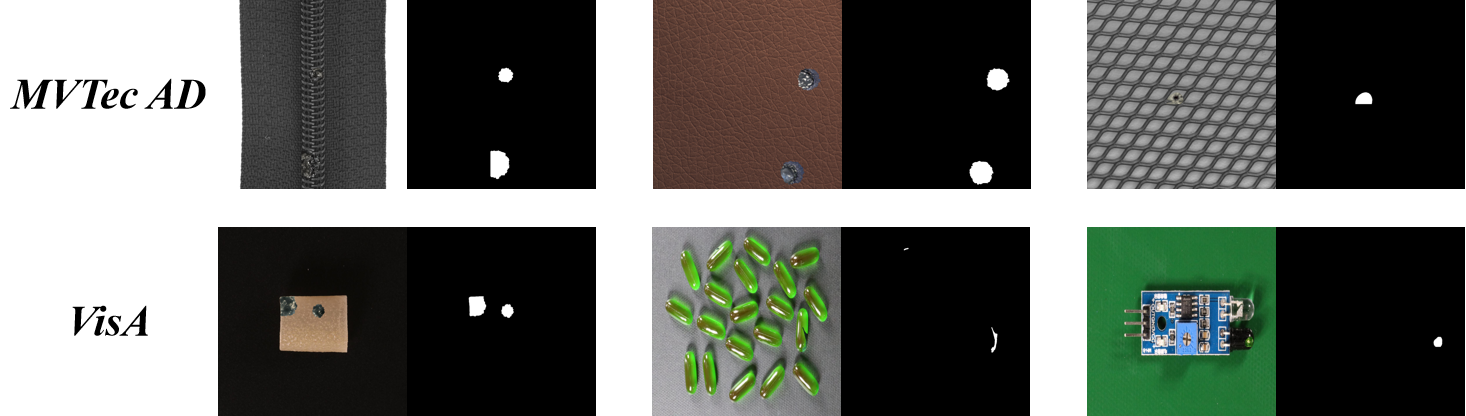}
        \caption{\textbf{\textit{ARAS}‑generated anomalies.}  Each shows: the locally edited image generated by \textit{ARAS} and binary anomaly mask. Rows correspond to \textit{MVTec~AD} (top) and \textit{VisA} (bottom) categories.}
        \label{fig:qual_syn}
    \end{subfigure}
    \begin{subfigure}[b]{0.95\linewidth}
        \centering
        \includegraphics[width=\linewidth]{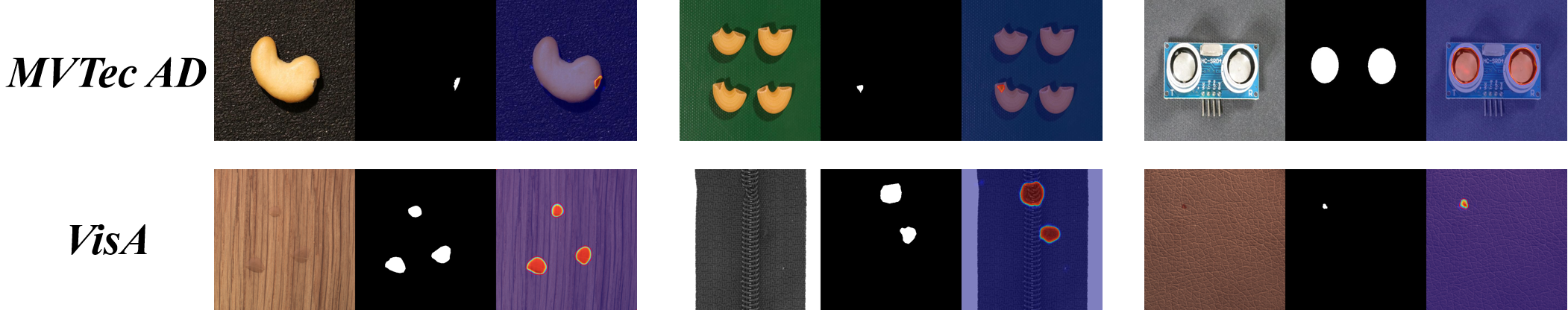}
        \caption{\textbf{\textit{QARAD} detection results.}  From left to right: defective image, ground‑truth mask, and \textit{QARAD} anomaly heatmap.}
        \label{fig:qual_res}
    \end{subfigure}
    \caption{Qualitative examples of (a) synthetic anomalies generated by \textit{ARAS} and (b) anomaly detection results produced by \textit{QARAD} on \textit{MVTec~AD} and \textit{VisA} datasets.}
    \label{fig:qual_visual}
\end{figure}

\begin{table*}[h]
\centering
\begin{tabular}{c|ccccccc}
\toprule
\textit{Category} &  \textit{DRAEM} 
&\textit{DSR} & \textit{CutPaste} & \textit{AnomalyDiffusion} & \textit{RD++}  & \textit{RealNet}
 & \textit{QARAD~(Ours)}
\\
\midrule
\textit{Bottle}     &  99.2/97.8 
&99.6/98.8 & \textbf{100}/99.1
& 99.8/99.4 & \textbf{100}/98.8  & \textbf{100}/99.3
 & \textbf{100}/\textbf{99.9}\\
\textit{Cable}      &  91.8/94.7 
&95.3/97.7 & 96.4/96.2
& \textbf{100}/99.2 & 99.3/98.4  &99.2/98.1
 &99.8/\textbf{99.8}\\
\textit{Capsule}    &  98.5/94.3 
&98.3/91.0 & 98.5/99.1
& 99.7/98.8& 99.0/98.8  &99.6/99.3
 &99.9/\textbf{99.9}\\
\textit{Hazelnut}   &  \textbf{100}/99.7 
&97.7/99.1 & \textbf{100}/99.0
& 99.8/99.8 & \textbf{100}/99.2  &\textbf{100}/99.8&99.8/\textbf{99.9}\\
\textit{Metal Nut}   &  98.7/99.5 
&99.1/94.1 & 99.9/98.0
& \textbf{100}/\textbf{99.8}& \textbf{100}/98.1  &99.7/98.6
 &99.8/\textbf{99.8}\\
\textit{Pill}       &  98.9/97.6 
&98.9/94.2 & 97.2/99.0
& 98.0/99.8 & 98.4/98.3  &99.1/99.0
 &\textbf{99.8}/\textbf{99.9}\\
\textit{Screw}      &  93.9/97.6 
&95.9/98.1 & 92.7/98.5
& 96.8/97.0 & \textbf{98.9}/\textbf{99.7}  &98.8/99.5
 &98.5/99.6\\
\textit{Toothbrush} &  \textbf{100}/98.1 
&\textbf{100}/\textbf{99.5} & 99.2/98.9
& \textbf{100}/99.2 & \textbf{100}/99.1  &99.4/98.7
 &99.5/\textbf{99.9}\\
\textit{Transistor} &  93.1/90.9 
&96.3/80.3 & 99.4/96.3
& \textbf{100}/99.3 & 98.5/94.3  &\textbf{100}/98.0 & 99.7/\textbf{99.8}\\
\textit{Zipper}     &  \textbf{100}/98.8 
&98.5/98.4 & 99.6/98.0
& 99.9/99.4 & 98.6/98.8  &99.8/99.2
 &99.7/\textbf{99.9}\\
\midrule
\textit{Carpet}     &  97.0/95.5 
&99.6/96.0 & 99.2/98.4
& 96.7/98.6 & \textbf{100}/99.2  &99.8/99.2
 &99.4/\textbf{99.8}\\
\textit{Grid}       &  99.9/\textbf{99.7} 
&\textbf{100}/99.6 & \textbf{100}/99.2
& 98.4/98.3 & \textbf{100}/99.3  &\textbf{100}/99.5&99.6/99.6\\
\textit{Leather}    &  \textbf{100}/98.6 
&99.3/99.5 & \textbf{100}/99.4
& \textbf{100}/99.8& \textbf{100}/99.5  &\textbf{100}/99.8&99.7/\textbf{99.9}\\
\textit{Tile}       &  99.6/99.2 
&\textbf{100}/98.6 & 99.9/97.6
& \textbf{100}/99.2 & 99.7/96.6  &99.9/99.4
 &  99.7/\textbf{99.8}\\
\textit{Wood}       &  99.1/96.4 
&94.7/91.5 & 99.0/95.0
& 98.4/98.9 & 99.3/95.8  &99.2/98.2
 &\textbf{99.9}/\textbf{99.8}\\
\midrule
\rowcolor[gray]{0.9}
\textbf{\textit{Avg.}}   &  98.0/97.2 &98.2/95.8 & 98.7/98.1
& 98.5/97.7 & 99.4/98.3  &99.6/99.0&\textbf{99.7}/\textbf{99.8}\\
\bottomrule
\end{tabular}

\caption{Performance comparison across different SOTA methods on \textit{MVTec~AD} dataset.}
\label{tab:mvtec}
\end{table*}

\begin{table*}[h]
\centering
\begin{tabular}{c|cccccc}
\toprule
\textit{Category}& \textit{SimpleNet}& \textit{SPADE}& \textit{RD} &\textit{RD++}& \textit{RealNet} & \textit{QARAD~(Ours)}
\\
\midrule
\textit{01} & 96.4/90.3& 91.4/97.3& 97.9/98.3 &96.8/96.2& \textbf{100}/98.2& \textbf{100}/\textbf{98.5}
\\
\textit{02} & 75.2/48.9& 71.4/94.4& 86.0/96.2 &90.1/96.4& 88.6/96.3&\textbf{90.5}/\textbf{96.7}
\\
\textit{03} & 99.3/97.2& 99.9/99.1& 99.7/94.2 &\textbf{100}/\textbf{99.7}& 99.6/99.4&99.5/98.9
\\
\midrule
\rowcolor[gray]{0.9}
\textbf{\textit{Avg.}}& 90.3/78.8& 87.6/96.9 & 94.5/96.2  &95.6/97.4& 96.1/97.9 & \textbf{96.7}/\textbf{98.0}
\\
\bottomrule
\end{tabular}

\caption{Performance comparison across different SOTA methods on \textit{BTAD} dataset.}
\label{tab:btad}
\end{table*}

Tab.~\ref{tab:visa}, ~\ref{tab:mvtec}, ~\ref{tab:btad} benchmark our \textit{QARAD} method against the most competitive methods on \emph{VisA}, \emph{MVTec~AD} and \emph{BTAD} datasets. Our \textit{QARAD} achieves new state-of-the-art performance in the most of all individual categories and improves the \textit{dataset‐level mean} to \textbf{98.9/99.8} (\textit{VisA}), \textbf{99.7/99.8} (\textit{MVTec~AD}), and \textbf{96.7/98.0} (\textit{BTAD}) in \textit{image‑ / pixel‑level AUROC} respectively surpassing the previous SOTA methods by \textit{+1.1/+1.0 points on VisA dataset}, \textit{+0.1/+0.8 points on MVTec~AD dataset} and \textit{+0.6/+0.1 points on BTAD dataset}.

The largest gaps appear on highly textured or fine‑grained classes such as \textit{Fryum} (+2.3/+3.7) and \textit{Wood}(+0.7/+1.6), corroborating that \textit{ARAS}’s token anchoring preserves micro‑structure that traditional pipelines cannot reconstruct. These categories contain filament‑scale fibres or glossy coating patterns whose local phase is easily destroyed by the noise‑re‑encode–upsample cycle of diffusion and the rigid paste geometry of template methods. Because \textit{ARAS} freezes all \textit{out‑of‑mask} \textit{VQ‑tokens} by our \textit{Masked AR Kernel}, the spectral statistics of the surrounding material flow seamlessly into the resampled region, preventing the artifacts that confuse our \textit{QARAD} detectors. On the \textit{BTAD} dataset, our \textit{QARAD} also outperform the previous SOTA methods. 

The aggregate evidence demonstrates that our pipeline delivers a \emph{holistic advance}: the training‑free \textit{ARAS} editor provides structurally consistent anomalies with linguistically controllable properties, while the quality‑aware re-weighting training scheme \textit{QAW} in \textit{QARAD} converts that realism into consistent performance gains over a spectrum of object geometries, texture regimes, and defect scales.

Representative qualitative results of \textit{ARAS}‑generated anomalies and \textit{QARAD} detection results, are provided in Fig.~\ref{fig:qual_visual}; the complete qualitative results, along with the raw input prompts fed to \textsc{Qwen model} and the textual descriptions it produces, appears in the \textit{Appendix}.

\subsection{Efficiency \& Computational Complexity}
\label{ssec:efficiency}

\begin{table}[h]
  \centering
    \resizebox{\linewidth}{!}{
      \begin{tabular}{lc}
        \toprule
        \textbf{Model} & \textbf{Inference Time (s/it)} \\
        \midrule
        \textit{Diffusion-based} (\eg \textit{RealNet's SDAS}) & 7.51 \\
        \rowcolor[gray]{0.9}
        \textbf{\textit{ARAS} (Ours)}                  & 1.49 \\
        \bottomrule
      \end{tabular}
    }
    \caption{Anomaly synthesis inference speed}
    \label{tab:anomaly_synthesis_speed}
\end{table}
\begin{table}[h]
    \centering
    \resizebox{\linewidth}{!}{
      \begin{tabular}{lcc}
        \toprule
        \textbf{Metric} & \textit{QARAD (w/o. QAW)} & \textbf{\textit{QARAD} (ours)} \\
        \midrule
        Train Time (s/ep)            & 100.89            & 110.36 \;(\,+9.47)   \\
        Inference Time (s/it)        & 0.21              & 0.21                \\
        \#Learnable Params (M)      & 524.11            & 524.11              \\
        \#Total Params (M)          & 590.94            & 742.22 \;(\,+151.28)\\
        \bottomrule
      \end{tabular}
    }
    \caption{Anomaly detection training \& inference complexity}
    \label{tab:anomaly_detection_complexity}
\end{table}

Tab.~\ref{tab:anomaly_synthesis_speed} reveals that our \textit{training‑free} \textit{ARAS} editor synthesises a masked anomaly $5.0\times$ faster than the diffusion‑based \textit{SDAS} pipeline used in \textit{RealNet} (1.49s~\emph{vs.}\ 7.51s per image at $1024^{2}$ resolution).  This speed‑up stems directly from the linear–token sampling rule in Eq.~\eqref{eq:masked_kernel}: runtime scales with the mask size only, whereas diffusion must iterate through at least 50–100 denoising steps irrespective of the edited area.

Turning to anomaly detection, Tab.~\ref{tab:anomaly_detection_complexity} compares plain \textit{RealNet} fine‑tuning with our full \textit{QARAD}.  Integrating the \textit{CLIP} backbone for image–text similarity scoring enlarges the total parameter budget by \textit{151M}; however, these parameters are \emph{frozen}, so the number of learnable weights remains unchanged \textit{(524M)}. Consequently, the training time per epoch increases only slightly by \textit{9.4s} (approximately \textit{9\%}), while the inference latency remains unchanged at \textit{0.21s} per image, as the \textit{QAW} module is exclusively utilized during training.

In aggregate, the proposed pipeline maintains \textit{RealNet}’s real‑time detection speed while accelerating anomaly synthesis by a factor of five, which is an attractive trade‑off for industrial scenarios requiring both rapid data augmentation and swift deployment, without compromising performance.

\subsection{Ablation Study}
\label{ssec:ablation}

\begin{table}[h]
    \centering
        \centering
        \resizebox{\linewidth}{!}{
            \begin{tabular}{lccc}
                \toprule
                \textbf{Method Variant} & \textbf{Image-AUROC} & \textbf{Pixel-AUROC} & \textbf{Comments}\\
                \midrule
                \textit{DREAM~(DTD)}& 98.0 & 97.2 & Original Synthesis Method \\
                \rowcolor[gray]{0.9}
                \textit{DREAM~(ARSR)}& 98.6 \textbf{\textit{(+0.6)}} & 97.9 \textbf{\textit{(+0.7)}} & Replaced with our \textit{ARAS} \\
                \midrule
                \textit{RealNet~(SDAS)}& 99.6 & 99.0 & Original Synthesis Method \\
                \rowcolor[gray]{0.9}
                \textit{RealNet~(ARSR)}& 99.6 \textbf{\textit{(+0.0)}} & 99.3 \textbf{\textit{(+0.3)}} & Replaced with our \textit{ARAS} \\
                \bottomrule
            \end{tabular}
        }
        \caption{Ablation of \textit{ARAS}.}
        \label{tab:ablation_aras}
\end{table}

\begin{table}[h]
        \centering
        \resizebox{\linewidth}{!}{
            \begin{tabular}{lccc}
                \toprule
                \textbf{\textit{QAW} Variant} & \textbf{Image-AUROC} & \textbf{Pixel-AUROC} & \textbf{Comments}\\
                \midrule
                \textit{w/o. QAW} & 99.6 & 99.3 & Uniform Weight\\
                \textit{Hinge QAW} & 99.6\textbf{\textit{(+0.0)}} & 99.8\textbf{\textit{(+0.5)}} & Clipped Linear Scaling \\
                \rowcolor[gray]{0.9}
                \textit{Softmax QAW} & 99.7\textbf{\textit{(+0.1)}} & \textbf{\textit{99.9(+0.6)}} & Softmax similarity weight\\
                \bottomrule
            \end{tabular}
        }
        \caption{Ablation of \textit{QAW}.}
        \label{tab:ablation_qaw}
\end{table}

Tab.~\ref{tab:ablation_aras}, ~\ref{tab:ablation_qaw} disentangles the respective contributions of our anomaly synthesis method \textit{ARAS} and our anomaly detection model \textit{QARAD}'s training scheme \textit{QAW}. All the ablation studies are experimented on the \textit{MVTec~AD} dataset.

\paragraph{Impact of \textit{ARAS}.}  
Replacing the augmentation-based and diffusion-inpainting pipelines in two representative baselines—\emph{DRAEM} (originally \textit{Perlin} masks with \textit{DTD} texture blending) and \emph{RealNet} (originally the \textit{SDAS} synthesis method)—with our training-free, token-anchored and language-conditioned editor \textit{ARAS} produces consistent gains: \textit{+0.6 image-AUROC} / \textit{+0.7 pixel-AUROC} for \textit{DRAEM}, and \textit{+0.3 pixel-AUROC} for \textit{RealNet} (Tab.~\ref{tab:ablation_aras}). The larger margin on \textit{DRAEM} is expected because its auto-encoder backbone is highly susceptible to high-frequency seam artefacts that \textit{ARAS} removes; by contrast, \textit{RealNet} already achieves \textit{image-level AUROC} very close to 100, so a modest additional improvement is both reasonable and informative. Qualitatively, \textit{ARAS} yields synthetic defects whose spectral statistics align with those of the surrounding material, thereby suppressing spurious responses along texture boundaries and producing cleaner heat-maps.

\paragraph{Impact of \textit{QAW}.}  
Starting from \textit{RealNet}~+~\textit{ARAS}, we compare three weighting schemes (Tab.~\ref{tab:ablation_qaw}). Uniform weighting (\textit{w/o. QAW}) serves as baseline. A \emph{hinge} mapping that truncates low‑similarity samples already lifts \textit{pixel‑AUROC} by \textit{0.5}, indicating that text‑image inconsistent anomalies indeed harm optimization. \emph{Softmax} calibration delivers the best performance (\textit{99.7 image-~/~99.9 pixel-level AUROC}), demonstrating that \emph{continuous attenuation} of moderate‑quality samples is superior to hard rejection, preserving diversity while still focusing gradients on high‑fidelity data. \textit{Appendix} will offer a complete theoretical analysis.

\paragraph{Orthogonality of the two components.}  
Because \textit{ARAS} and \textit{QAW} operate at distinct stages (data generation vs.\ loss re‑weighting), their gains accumulate almost additively: the full \textit{QARAD} configuration outperforms the \textit{diffusion}-based by \textit{+0.1 image-level AUROC~/~+0.6 pixel-level AUROC} on the \textit{MVTec~AD} dataset without increasing inference latency. This validates our design hypothesis that high‑quality local edits and quality‑aware training are mutually reinforcing.

\section{Conclusions}
\label{sec:conclusion}

Our work presents a fully anomaly synthesis and detection pipeline for industrial anomaly detection that integrates two novel components: (i) \textbf{\textit{ARAS}}, a token‑anchored auto‑regressive editor that inserts linguistically controlled, micro‑structure‑preserving defects without re‑encoding the surrounding image context, and (ii) \textbf{\textit{QARAD}}, a quality‑aware re-weighting training scheme that leverages image–text similarity to modulate the influence of each synthetic sample during detector optimization. Extensive experiments on three public benchmarks, \emph{MVTec~AD}, \emph{VisA}, and \emph{BTAD}, demonstrate that our method establishes new SOTA performance at both \textit{image-} and \textit{pixel-level AUROC}, while reducing anomaly synthesis latency by a factor of five relative to diffusion‑based approaches. Ablation studies confirm that \textit{ARAS} and \textit{QARAD} contribute complementary gains, with \textit{softmax} calibration yielding the most stable and accurate results. Beyond accuracy and efficiency, \textit{ARAS} delivers fine‑grained semantic controllability, enabling users to steer defect attributes (size, orientation, material phase) via natural language. Representative qualitative results highlight sharper, seam‑free edits and cleaner detection heatmaps, reinforcing the quantitative findings. In addition, we will publicly release the large-scale image–mask–text dataset generated by our \textit{ARAS}. Overall, our study shows that high‑fidelity local synthesis \emph{plus} principled sample re‑weighting can significantly narrow the realism gap in synthetic anomaly data, paving the way for faster, more reliable industrial anomaly detection systems.

{
    \small
    \bibliographystyle{ieeenat_fullname}
    \bibliography{main}
}

\clearpage
\setcounter{page}{1}
\maketitlesupplementary
\renewcommand\thesection{\Alph{section}}
\renewcommand\thesubsection{\Alph{section}.\arabic{subsection}}
\setcounter{section}{0}
\setcounter{subsection}{0}
%———图（figure）———
\setcounter{figure}{0}          % 计数器归零
\renewcommand{\thefigure}{S\arabic{figure}} % 定义为 Fig.S1、Fig.S2 …

%———表（table）———
\setcounter{table}{0}
\renewcommand{\thetable}{S\arabic{table}}   % 定义为 Table S1、Table S2 …

% 如果还用到了子图/子表（subfigure, subtable）：
\captionsetup[subfigure]{labelformat=simple,labelsep=space}
\renewcommand\thesubfigure{\thefigure\alph{subfigure}}  % 例如 Fig.S1a、Fig.S1b
\captionsetup[subtable]{labelformat=simple,labelsep=space}
\renewcommand\thesubtable{\thetable\alph{subtable}}

\section{Reproducibility Hyper-parameter List}
The following tables ~\ref{tab:aras_par}, ~\ref{tab:qarad_par} collect all tunable settings that must be exposed for exact reproduction of our results. Every experiment in the paper uses these values unless stated otherwise. All other options not shown here are kept exactly as in the public codebase and the original pape.

Tab.~\ref{tab:aras_par} lists our \textit{ARAS} anomaly synthesis method hyper-Parameter. Table.~\ref{tab:qarad_par} records our \textit{QARAD} anomaly detection method hyper-Parameter.

\section{Sensitivity to the soft-max temperature
  \texorpdfstring{$\gamma$}{gamma}}
  
To check the robustness of \textit{QARAD} under the soft-max variant \(w(s)=\exp(\gamma s)\), we trained the detector with saome different temperatures. Partial results on the \textit{MVTec~AD} dataset are summarized in tab.~\ref{tab:gamma-scan}.  Performance remains stable for \(\gamma\in[1,5]\); the default \(\gamma=2.5\) stays the best performance.

\begin{table}[h]
\centering
\begin{tabular}{c cc}
\toprule
$\gamma$ & \textit{Image-AUROC} & \textit{Pixel-AUROC}\\
\midrule
1 & 99.4 & 99.6\\
5 & 99.2 & 99.6\\
\rowcolor[gray]{0.9}
2.5 (default) & 99.7 & 99.9\\
\bottomrule
\end{tabular}
\caption{Partial Results on effect of soft-max temperature.}
\label{tab:gamma-scan}
\end{table}

\begin{table*}[htbp]
\centering
\begin{tabular}{c c l}
\hline
\textbf{Parameter} & \textbf{Default} & \textbf{Description}\\
\hline
\texttt{seed} & 123 & Base random seed.\\
\texttt{tau} & 0.7 & Self-attention temperature; lower values yield sharper attention.\\
\texttt{use\_bit\_label} & 1 & Use bit-wise labels for token prediction.\\
\texttt{sampling\_per\_bits} & 1 & Number of sampling attempts per bit during generation.\\
\texttt{gt\_leak} & -1 & Number of coarse scales leaked from ground truth during AR generation ($-1$ leaks all).\\
\hline
\end{tabular}
\caption{Core \textit{ARAS} hyper-parameters retained for reproducibility.}
\label{tab:aras_par}
\end{table*}

\begin{table*}[htbp]
\centering
\begin{tabular}{c c l}
\hline
\textbf{Parameter} & \textbf{Default} & \textbf{Description}\\
\hline
\texttt{seed} & 123 & Base random seed.\\
\texttt{ep} & 2000 & Total number of training epochs.\\
$\gamma$ (softmax) & 2.5 & Temperature in $w_{\text{soft}}(s)=\exp(\gamma s)$.\\
$\beta$ (hinge) & 0.20 & Threshold in $w_{\text{hinge}}(s)=\max{0,,s-\beta}$.\\
\hline
\end{tabular}
\caption{Core \textit{QARAD} hyper-parameters retained for reproducibility.}
\label{tab:qarad_par}
\end{table*}

\begin{table*}[htbp]
\centering
\begin{tabular}{@{}c p{15cm}@{}}
\toprule
\textbf{\textless{}category\textgreater{}} & \textbf{\textless{}category\_specific\_description\textgreater{}} \\ \midrule
\textit{bottle} & This is a top view of the bottle mouth, and the bottle appears to be made of glass with a circular, smoothly
rounded rim, a faint brownish tint around the inner edge, and a dark interior that looks nearly black due to
its depth, showcasing a uniformly thick, intact seal area. \\
\textit{cable} & This is a top-down cross-section of a multi-core cable showing three color-coded conductors (green-yellow,
blue, and beige) each containing tightly packed copper strands, all arranged neatly within a circular white
outer sheath. \\
\textit{capsule} & This is a two-toned gelatin capsule featuring a black half and an orange half marked with “500,” showing a
smooth, cylindrical form with rounded ends and clear, legible text. \\
\textit{carpet} & This is a close-up photograph of a woven, grayish carpet featuring a square-like interlacing of fibers that
creates a tightly knit, textured surface with a uniform cross pattern. \\
\textit{grid} & This is a close-up view of a diamond-shaped lattice grid made of interwoven lines arranged diagonally,
creating a uniform repeating pattern with evenly spaced openings through which a lighter background is
visible. \\
\textit{hazelnut} & This is a single hazelnut featuring a warm brown shell with subtle vertical ridges, a slightly rounded body that tapers toward the base, and a lighter, dome-like top with a rough, woody texture. \\
\textit{leather} & This is a close-up of a brown leather surface displaying a finely textured, pebbled grain pattern with subtle
creases and a uniformly warm tone across its slightly matte finish. \\
\textit{metalnut} & This is a top-down view of a four-lobed metal nut with each lobe curving outward in a clockwise arrangement
around the central threaded hole, featuring a frosted, translucent finish and a symmetrically balanced shape. \\
\textit{pill} & This is an oval, white compressed tablet pill with scattered red specks on its surface and the embossed
letters "FF" in the center, giving it a speckled, slightly textured appearance. \\
\textit{screw} & This is a metallic screw with a countersunk head featuring a cross-shaped drive, a slender cylindrical shaft,
and evenly spaced threads running toward a sharply pointed tip. \\
\textit{tile} & This is a close-up of a ceramic tile featuring an irregular distribution of dark speckles scattered across a
lighter background, creating a mottled, stone-like appearance throughout its glossy surface. \\
\textit{toothbrush} & This is a top-down view of a toothbrush head, featuring a slightly elongated oval shape with multiple rows of
densely packed, alternating white and other color bristles anchored in a white plastic base. \\
\textit{transistor} & This is a transistor in a black plastic case with three bent metal leads inserted into the middle three
adjacent holes in the bottom row of a copper prototyping board (each hole is surrounded by a copper pad),
demonstrating the standard TO-92 (or similar) package design. \\
\textit{wood} & This is a close-up of a wooden surface displaying parallel vertical grain lines in varying shades of warm
brown, creating a natural striped pattern across its smooth finish. \\
\textit{zipper} & This is a close-up view of a black zipper featuring two textured fabric tapes joined by a row of uniformly
interlocking plastic teeth down the center, creating a neat, continuous closure mechanism. \\
\bottomrule
\end{tabular}
\caption{Exact strings that replace the angle-bracket placeholders in the prompt on the \textit{MVTec~AD} dataset.}
\label{tab:placeholder-mvtec}
\end{table*}

\begin{table*}[htbp]
\centering
\begin{tabular}{@{}c p{15cm}@{}}
\toprule
\textbf{\textless{}category\textgreater{}} & \textbf{\textless{}category\_specific\_description\textgreater{}} \\ \midrule
\textit{candle} & This is a top-down view of four round, cream-colored tealight candles arranged in a neat 2 × 2 grid on a dark
background, each showing a smooth, circular wax surface with a short central wick and soft, warm lighting that
casts subtle shadows around the uniformly shaped discs. \\
\textit{capsules} & This is a top-down view of multiple glossy, translucent green soft-gel capsules—each elongated with gently
rounded ends and bright specular highlights—randomly scattered across a lightly textured gray surface. \\
\textit{cashew} & This is a top-down view of a single cashew nut displaying a curved, kidney-like silhouette in warm light
beige, its smooth matte surface showing subtle speckles and gentle contours, all set against a dark, coarse-
textured background that accentuates its distinctive shape. \\
\textit{chewinggum} & This is a top-down view of a single rectangular white chewing-gum pellet with softly rounded corners and a
smooth, slightly glossy coating, resting on a dark, coarse-textured background that starkly contrasts its
clean, uniform surface. \\
\textit{fryum} & This is a top-down view of a single pale-orange, wheel-shaped fryum snack with six evenly spaced radial spokes
converging at a small central hole, forming a thin crisp rim whose lightly textured, airy structure contrasts
against the dotted green background. \\
\textit{macaroni1} & This is a top-down view of four elbow-shaped macaroni pieces arranged in a neat 2 × 2 grid on a textured green
background, each warm orange tube showing smooth inner walls and evenly spaced exterior ridges along its
outward-curving arc. \\
\textit{macaroni2} & This is a top-down view of four pale-yellow elbow macaroni pieces arranged in a tidy 2 × 2 grid on a dotted
green background, each hollow arc displaying concentric exterior ridges and neatly cut ends that reveal a
smooth inner channel. \\
\textit{pcb1} & This is a top-down view of an HC-SR04 ultrasonic sensor module consisting of a small rectangular blue PCB
bearing two circular metal-rimmed transducers with fine mesh (receiver on the left, transmitter on the right),
a row of four vertical header pins labeled UCC, Trig, Echo, and GND protruding from the top edge, and small
surface components plus mounting holes at each corner. \\
\textit{pcb2} & This is a top-down view of the rear side of an HC-SR04 ultrasonic sensor module: a blue rectangular PCB
bearing three small SOIC integrated circuits (LM324M on the right, RCWL-9206 on the left, a driver in the
middle), two vertical rows of six SMD resistor networks flanking a central set of four through-hole header
pins (VCC, Trig, Echo, GND) that protrude upward from the board’s top edge, with white silkscreen labels,
visible copper traces, and a mounting hole at each corner. \\
\textit{pcb3} & This is a top-down view of a compact blue PCB IR sensor module with three horizontal header pins on the left
(VCC, GND, OUT), a central cluster of surface-mount resistors and an 8-pin IC, a blue trimmer potentiometer
marked “3362,” and—protruding from the right edge—a clear LED emitter paired with a black photodiode receiver,
all set against a textured green background. \\
\textit{pcb4} & This is a top-down view of a compact blue Li-ion charger module featuring a micro-USB input connector on the
left, white-silkscreen IN+/IN- pads at opposing corners, a cluster of tiny surface-mount resistors and
capacitors, an 8-pin controller IC labeled “4056E,” and clearly marked BAT+ and BAT- output pads on the right,
all centered on a rectangular PCB with four corner mounting holes. \\
\textit{pipe fryum} & This is a top-down view of a single beige pipe-shaped fryum—a short, hollow cylinder with cleanly cut straight
ends and a subtly rough, speckled surface—set against a deep black background that emphasizes its tubular
form. \\
\bottomrule
\end{tabular}
\caption{Exact strings that replace the angle-bracket placeholders in the prompt on the \textit{VisA} dataset.}
\label{tab:placeholder-visa}
\end{table*}

\begin{table*}[htbp]
\centering
\begin{tabular}{@{}c p{15cm}@{}}
\toprule
\textbf{\textless{}category\textgreater{}} & \textbf{\textless{}category\_specific\_description\textgreater{}} \\ \midrule
\textit{01} & This is a top view of a metallic annular part whose smooth, silver‑gray surface displays many evenly spaced
concentric machining grooves that form bright circular bands around a large central hole, all lying flat
against a dark background. \\
\textit{02} & This is a planar view of a rectangular board‑like surface coated with uniform, parallel vertical streaks that
look like fine wood‑grain striations, giving the whole beige panel a consistent striped texture. \\
\textit{03} & This is a top view of a round disc whose smooth turquoise face exhibits a series of concentric ridges that
create light‑to‑dark rings from center to rim, producing a perfectly symmetrical bull‑eye pattern. \\
\bottomrule
\end{tabular}
\caption{Exact strings that replace the angle-bracket placeholders in the prompt on the \textit{BTAD} dataset.}
\label{tab:placeholder-btad}
\end{table*} 

\section{Theoretical Analysis of Quality-Aware Weighting}
\label{app:theory}

We formalise why the \emph{softmax} weighting scheme is preferable to a
\emph{linear} (hinge-style) alternative.

%---------------------------------------------------------------------------
\subsection{Problem setup}

For each synthetic sample $(x_i,y_i,s_i)$ we have a similarity
score $s_i\!\in[-1,1]$ and loss
$\ell_i=\ell\!\bigl(f_\varphi(x_i),y_i\bigr)\!\ge0$.
Given any monotone increasing weight function $w(\cdot)$ define
\[
  \tilde w_i=\frac{w(s_i)}{\frac1N\sum_{j=1}^{N}w(s_j)},\quad
  \hat L_{w}=\frac1N\sum_{i=1}^{N}\tilde w_i\,\ell_i.
  \label{eq:weighted-risk}
\]
The two concrete choices we compare are
\[
  \textit{Hinge: } w_{\text{lin}}(s)=\max\{0,s-\beta\},\;
  \beta\le1,
\]
\[
  \textit{Softmax: } w_{\text{soft}}(s)=e^{\gamma s},\;
  \gamma>0.
\]

\noindent \textbf{Standing assumptions.}

(A1) $\ell(\theta;x)$ is $L$-Lipschitz in $x$;  

(A2) $\ell_i$ is $\sigma_0$-sub-Gaussian;  

(A3) $\operatorname{Cov}(\ell_i,s_i)\le0$ (empirically true because
higher image–text consistency implies lower loss).

%---------------------------------------------------------------------------
\subsection{Unbiasedness}

Using importance-sampling algebra one checks that
\[
  \mathbb E[\hat L_{w}]
  =\frac1N\sum_{i=1}^{N}
    \mathbb E\!\Bigl[
      \frac{w(s_i)}{\tfrac1N\sum_j w(s_j)}\,\ell_i
    \Bigr]
  =\mathbb E_{p_{\text{high}}}[\ell],
\]
so \(\hat L_{w}\) is an unbiased estimate of the
high-quality population risk with density
\(p_{\text{high}}(x)\propto p(x)w\bigl(s(x)\bigr)\).

%---------------------------------------------------------------------------
\subsection{Why the variance of softmax is minimal}

Minimising the second moment
\(\sigma^{2}(w)=\operatorname{Var}(\tilde w_i\ell_i)\)
subject to \(\sum_i\tilde w_i=1\) and an entropy floor
\(-\sum_i\tilde w_i\log\tilde w_i\ge H_0\)
leads—via the usual Lagrange-multiplier calculus—to the exponential-family
solution
\[
  \tilde w_i^\star\propto\exp\bigl(\lambda_1\ell_i+\lambda_2\bigr).
\]
Because \(\ell_i\) is (approximately) affine in \(s_i\)
under assumption (A3),
this reduces to the softmax form
\(w^\star(s)\propto e^{\gamma s}\).
Hence softmax gives the \emph{variance-optimal} weights
under an entropy constraint.

%---------------------------------------------------------------------------
\subsection{Softmax \emph{strictly} lowers variance relative to linear}

Let
\(\sigma^2(\gamma)=\operatorname{Var}\!\bigl(\tilde w_{\text{soft}}(\gamma)
  \,\ell\bigr)\) with
\(\tilde w_{\text{soft}}(\gamma)\propto e^{\gamma s}\).
Taylor expansion at \(\gamma=0\) yields
\[
  \frac{\partial\sigma^{2}}{\partial\gamma}\bigl|_{\gamma=0}
  =2\,\operatorname{Cov}(\ell,s)
  < 0\quad\text{by (A3).}
\]
Thus for every small positive \(\gamma\) we have
\(\sigma^{2}(\gamma)<\sigma^{2}(0)\),
and because linear weighting is exactly the first-order (hinge) truncation
of \(e^{\gamma s}\), there exists a
\(\gamma^\star>0\) such that
\[
  \sigma^{2}(\gamma^\star)\le
  \operatorname{Var}\!\bigl(\tilde w_{\text{lin}}\ell\bigr).
\]
In words, \emph{softmax never increases and typically decreases the
gradient variance relative to a linear/hinge rule}.

%---------------------------------------------------------------------------
\subsection{Implication for generalisation}

Catoni-style PAC-Bayes gives(constants and KL term suppressed for brevity), with probability at least \(1-\delta\),
\[
  \mathcal E(\hat f)-\mathcal E^\star
  \le
  \sqrt{\frac{2\,\sigma^{2}(w)\,\ln(1/\delta)}{N}}.
\]
Since \(\sigma^{2}(w_{\text{soft}})\le\sigma^{2}(w_{\text{lin}})\),
the generalization bound is strictly tighter for softmax weighting,
explaining the AUROC gain reported in the main paper.

Softmax weighting is unbiased, variance-optimal under an entropy constraint, guarantees lower (or equal) variance than linear weighting, and therefore enjoys a provably tighter PAC-Bayes generalisation bound. These facts fully justify its use.

\section{Prompt used for synthetic‐defect generation}

The complete system + user prompt that we feed into
\textsc{Qwen 2.5-VL-32B-Instruct} is reproduced below.
Angle-bracket tokens mark runtime substitutions:
\textless{}category\textgreater{} is the current object class
(“bottle”, “capsule”, …);
\textless{}category\_specific\_description\textgreater{} is its pre-written
domain note;
\textless{}image\textgreater{} and \textless{}mask\textgreater{} are the user-supplied
original RGB image and the corresponding binary anomaly mask from
\textit{MaPhC2F}, respectively.
Complete \textless{}category\textgreater{}–\textless{}category\_specific\_description\textgreater{}
pairs can be seen in~\ref{tab:placeholder-mvtec},
~\ref{tab:placeholder-visa}, and~\ref{tab:placeholder-btad}.

\begin{quote}\begingroup\footnotesize\ttfamily
\#\#\# system  

                You are a vision expert who describes possible anomalies of an object and can introduce possible anomalies into a certain area of the normal image provided by the user and an anomaly mask showing the location of the anomaly in the image.
                
                Instructions:
                
                    - You will receive images of an object along with a binary mask. The white pixels in the mask represent the anomaly area of the object.
                    
                    - The current object category is <category>. <category\_specific\_description>
                    
                    - Imagine and describe realistically DEFECT that might appear on this category of the defect type and match the image and anomaly mask provided. Only randomness and fit is considered, no other considerations are needed.
                    
                    - A step-by-step analysis that clearly describes the logical process for identifying where the defect may be, its size relative to the object, its visual characteristics, and its approximate coordinate range in the image.
                    
                    - A clear description of the shape and color of the defect, explaining what unique shape characteristics and color it has.
                    
                    - Please watch the image provided carefully because the image provided may be different from the one you are familiar with.
                    
                    - Output your analysis strictly in the following structured format and order in English:
                    
                    Reasoning:
                    
                        1. [First step in your logical reasoning process]
                        
                        2. [Second step in your logical reasoning process]
                        
                        3. [Third step in your logical reasoning process]
                        
                    Description:
                    
                        [A concise and clear description of the single identified defect, specifying location, appearance, color, shape description, coordinate range, and approximate size relative to the object. less than 35 tokens.   Must mention: type / size / color-tone / shape]
                        
                    Parser:
                    
                        type: [selected defect type from the provided list]
                        
                        location: [specific location on the object where the defect is observed]
                        
                        coordinate: [coordinate range in terms of proportions of the image, e.g., "from 1/4 to 3/4 horizontally, and from 1/2 to top edge vertically"]
                        
                        size: [estimated size as a percentage of the object or area affected]
                        
                        shape: [a brief description of the defect's shape characteristics, such as "an irregular ellipse" or "elongated irregular shape with indentations and protrusions"]
                        
                        SupplementaryMaterials: [A brief summary of the visual features and any additional information not previously mentioned]

\#\#\# user  

Original Image: <image> Anomaly Mask From MaPhC2F dataset: <mask>  

                Please carefully examine the provided image and its anomaly mask because the image provided may be different from the one you are familiar with. <category\_specific\_description>
                
                Describe ONE realistic defect that could occur on this <category> and match the image and anomaly mask provided.
                
                Remember to choose the defect type RANDOMLY(Only randomness and fit is considered, no other considerations are needed).
                
                Follow strictly the Reasoning, Description, and Parser format as instructed in English. Description must be concise and clear, less than 35 tokens, and must mention: type / size / color-tone / shape.
                
\endgroup\end{quote}

\section{Key Pseudocode for Reproducibility}

The complete data flow consists of three clearly important stages:

(i) \emph{building image-mask-prompt triplets},

(ii) \emph{ARAS} anomaly synthesis, and

(iii) \emph{QARAD} anomaly detection.  

Algorithms\,\ref{alg:preproc}, \ref{alg:mar-kernel}, \ref{alg:qarad-rev} record the exact procedures used in our codebase.

\begin{itemize}
\item \textbf{Pre-processing \& prompt builder}  
      Reads every normal image, resizes it and its binary mask to $1024\times1024$, fills the system / user template with the category-specific sentence, and stores the tokenised chat in JSON—this is exactly what \textsc{Qwen 2.5-VL-32B-Instruct} consumes at generation time.

\item \textbf{\textit{ARAS}: anomaly synthesis}  
        A frozen auto-regressive VQ decoder edits \emph{only} the tokens
        addressed by the anomaly mask and keeps the surrounding lattice intact.
        Algorithm \ref{alg:mar-kernel} is an explicit construction of the
        token-anchored kernel \(K_{\tau,C}\) defined by Eq.\,(3):
        every context token in \(C\) is hard–gated via
        \(\mathrm{Gate}_C\!\bigl(\,\cdot\,\bigr)\) (Eq.\,(2)),
        while the masked set \(M\) is visited once in a random order~\(\pi\).
        
\item \textbf{\textit{QARAD} training loop}  
      Synthetic triplets \((\hat{x},m,\tau)\) generated by ARAS vary in
        semantic fidelity.  QARAD down-weights low-consistency outliers with a
        monotone calibration \(\varphi(s)\):
        soft-max \(\varphi_{\text{soft}}(s)=\exp(\gamma s)\) or
        hinge \(\varphi_{\text{hinge}}(s)=\max\{0,\,s-\beta\}\).
        Algorithm \ref{alg:qarad-rev} implements the resulting weighted-risk
        training loop used for all detector experiments.
\end{itemize}

\begin{algorithm}[H]
\caption{Dataset pre-processing and prompt construction}
\label{alg:preproc}
\begin{algorithmic}[1]
\Require Root directory $D$, category list $\mathcal{C}$,
         refined masks $M$ (MaPhC2F),
         Qwen\,2.5 processor \texttt{proc},
         template strings \texttt{SYS}, \texttt{USR}
         
\ForAll{$c \in \mathcal{C}$}
   \State Read description file $d_c$
   \State Gather good images in $D$
   \State Sample k mask
   \ForAll{image $x_i$}
      \State Resize $x_i$ and mask $m_c(x_i)$ to $1024 \times 1024$
      \State Substitute $\langle$category$\rangle \!\gets c$ and
             $\langle$category\_specific\_description$\rangle \!\gets d_c$
             in \texttt{SYS}, \texttt{USR}
      \State Build chat \textit{messages} $=\{\texttt{SYS},\texttt{USR},x_i,m_c(x_i)\}$
      \State Encode with \texttt{proc} and write JSON
   \EndFor
\EndFor
\end{algorithmic}
\end{algorithm}

\begin{algorithm}[H]
\caption{Token-anchored masked-AR sampling ($K_{\tau,C}$)}
\label{alg:mar-kernel}
\begin{algorithmic}[1]
\Require normal image $x$; binary mask $m$ ($m_{ij}\in\{0,1\}$); prompt $\tau$
\State $z \gets \mathrm{VQ}\text{-}\mathrm{Encode}(x)$
\State $C \gets \{(i,j)\mid m_{ij}=0\}$,\; $M \gets \{(i,j)\mid m_{ij}=1\}$
\State draw a random bijection $\pi:\{1,\dots,|M|\}\to M$
\For{$t=1$ \textbf{to} $|M|$}
    \State $(i,j) \gets \pi(t)$
    \State $q \gets p_\theta\bigl(z_{\pi(<t)} \,\big|\, z_C,\tau\bigr)$
    \State $q \gets \mathrm{Gate}_C(q)$   \Comment{force unit prob.\ on $z_C$}
    \State sample $z_{ij} \sim q_{ij}$ and write back
\EndFor
\State $\hat{x} \gets \mathrm{VQ}\text{-}\mathrm{Decode}(z)$
\State \Return $(\hat{x}, m, \tau)$
\end{algorithmic}
\end{algorithm}

\begin{algorithm}[H]
\caption{QARAD detector optimisation}
\label{alg:qarad-rev}
\begin{algorithmic}[1]
\Require mini-batch $\mathcal{B}=\{(\hat{x}_i,m_i,\tau_i)\}_{i=1}^B$;
         detector $f_\theta$; pixel loss $\ell$;
         calibration $\varphi_{\text{soft}}(\gamma)$ or $\varphi_{\text{hinge}}(\beta)$
\ForAll{$i=1\ldots B$}
    \State $s_i \gets \langle g_{\text{img}}(\hat{x}_i),\,g_{\text{txt}}(\tau_i)\rangle$  \Comment{CLIP cosine}
    \State $w_i \gets \varphi(s_i)$
\EndFor
\State normalise $w_i \gets w_i / \sum_{j} w_j$
\State $\mathcal{L} \gets \sum_{i} w_i\,\ell\!\bigl(f_\theta(\hat{x}_i),\,m_i\bigr)$
\State update $\theta \gets \theta - \eta \nabla_\theta \mathcal{L}$
\end{algorithmic}
\end{algorithm}

\section{Qualitative Results}
To intuitively demonstrate the effectiveness of our proposed methods, we present qualitative results generated by the \textit{ARAS} anomaly synthesis method and our anomaly detection approach, \textit{QARAD}, across multiple datasets. We visually illustrate anomalies created by anomaly mask and textual prompts, as well as detection results compared against the ground-truth masks. These visualizations highlight the precision, robustness, and generalization capabilities of our methods in realistic anomaly scenarios.

\subsection{Qualitative Results of \textit{ARAS}-generated Anomalies}
\begin{figure*}
    \centering
    \includegraphics[width=1\linewidth]{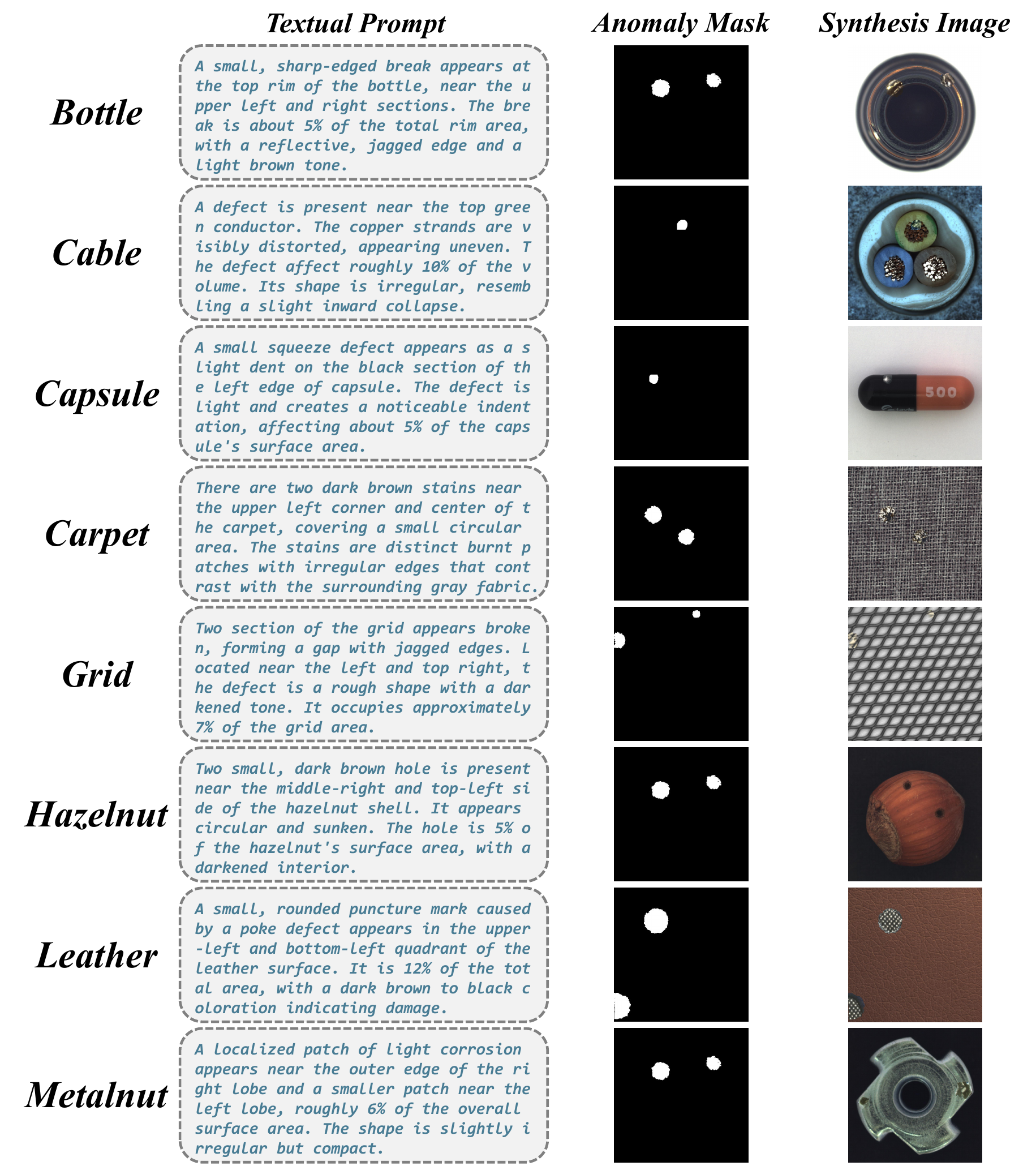}
    \caption{Qualitative results of \textit{ARAS}-generated anomalies and the corresponding textual prompts and anomaly masks on the \textit{MVTec~AD} dataset. (Part \textit{I})}
    \label{fig:mvtec_sys_1}
\end{figure*}

\begin{figure*}
    \centering
    \includegraphics[width=1\linewidth]{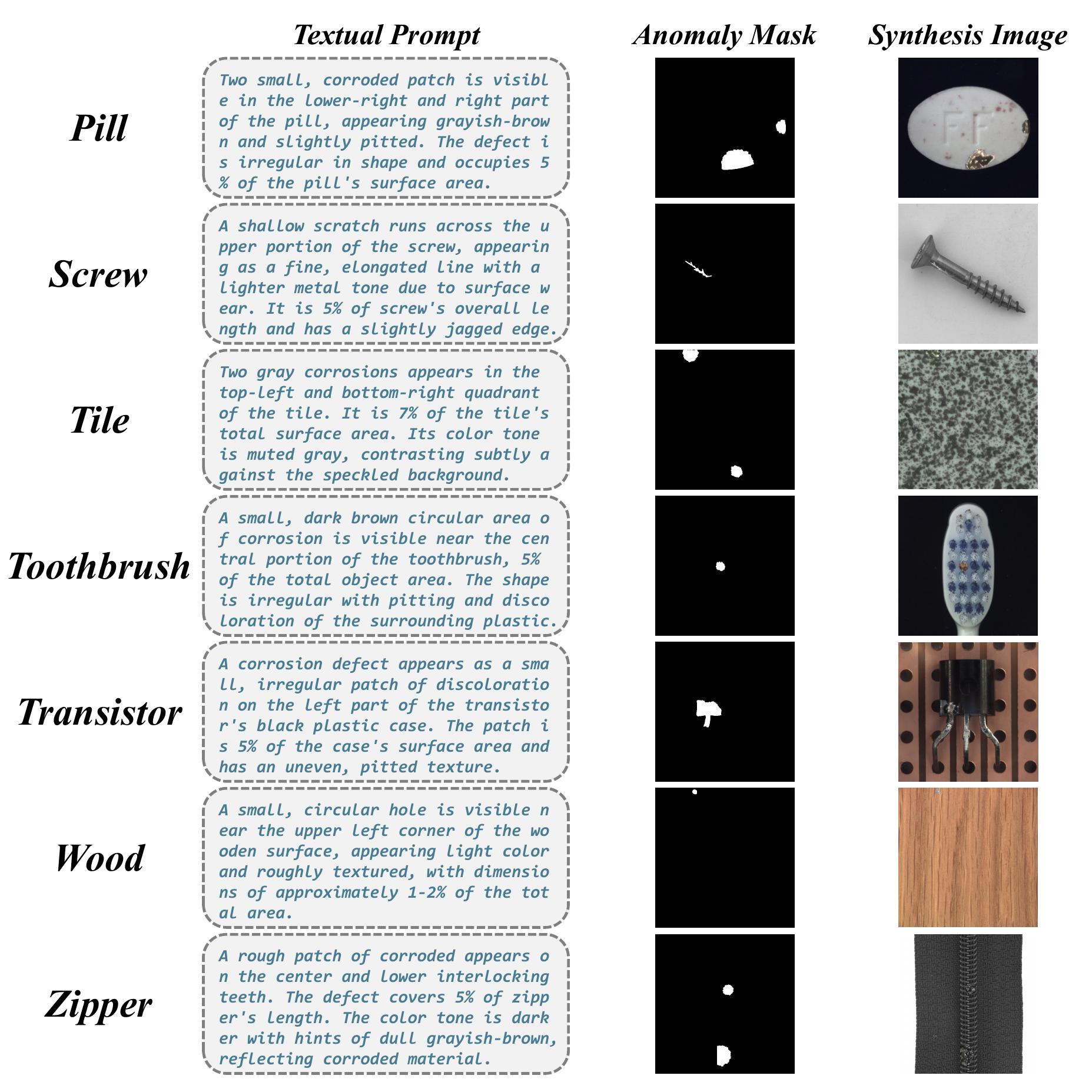}
    \caption{Qualitative results of \textit{ARAS}-generated anomalies and the corresponding textual prompts and anomaly masks on the \textit{MVTec~AD} dataset. (Part \textit{II})}
    \label{fig:mvtec_sys_2}
\end{figure*}

Fig.~\ref{fig:mvtec_sys_1} and \ref{fig:mvtec_sys_2} present synthesized anomalies on the widely used industrial anomaly detection dataset, \textit{MVTec~AD}. The anomalies clearly reflect the textual prompts, exhibiting diverse and realistic anomaly patterns. 

\begin{figure*}
    \centering
    \includegraphics[width=1\linewidth]{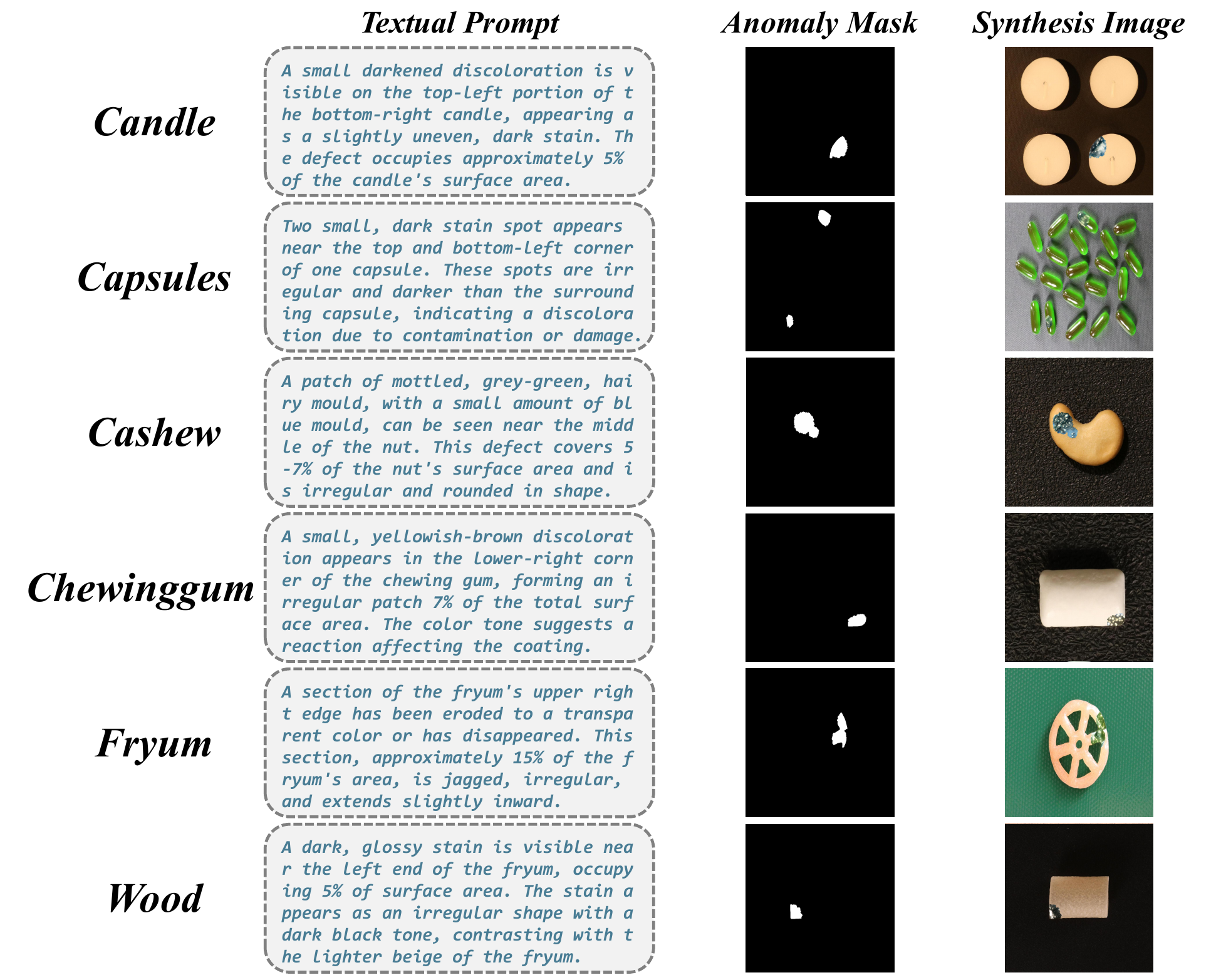}
    \caption{Qualitative results of \textit{ARAS}-generated anomalies and the corresponding textual prompts and anomaly masks on the \textit{VisA} dataset. (Part \textit{I})}
    \label{fig:visa_sys_1}
\end{figure*}

\begin{figure*}
    \centering
    \includegraphics[width=1\linewidth]{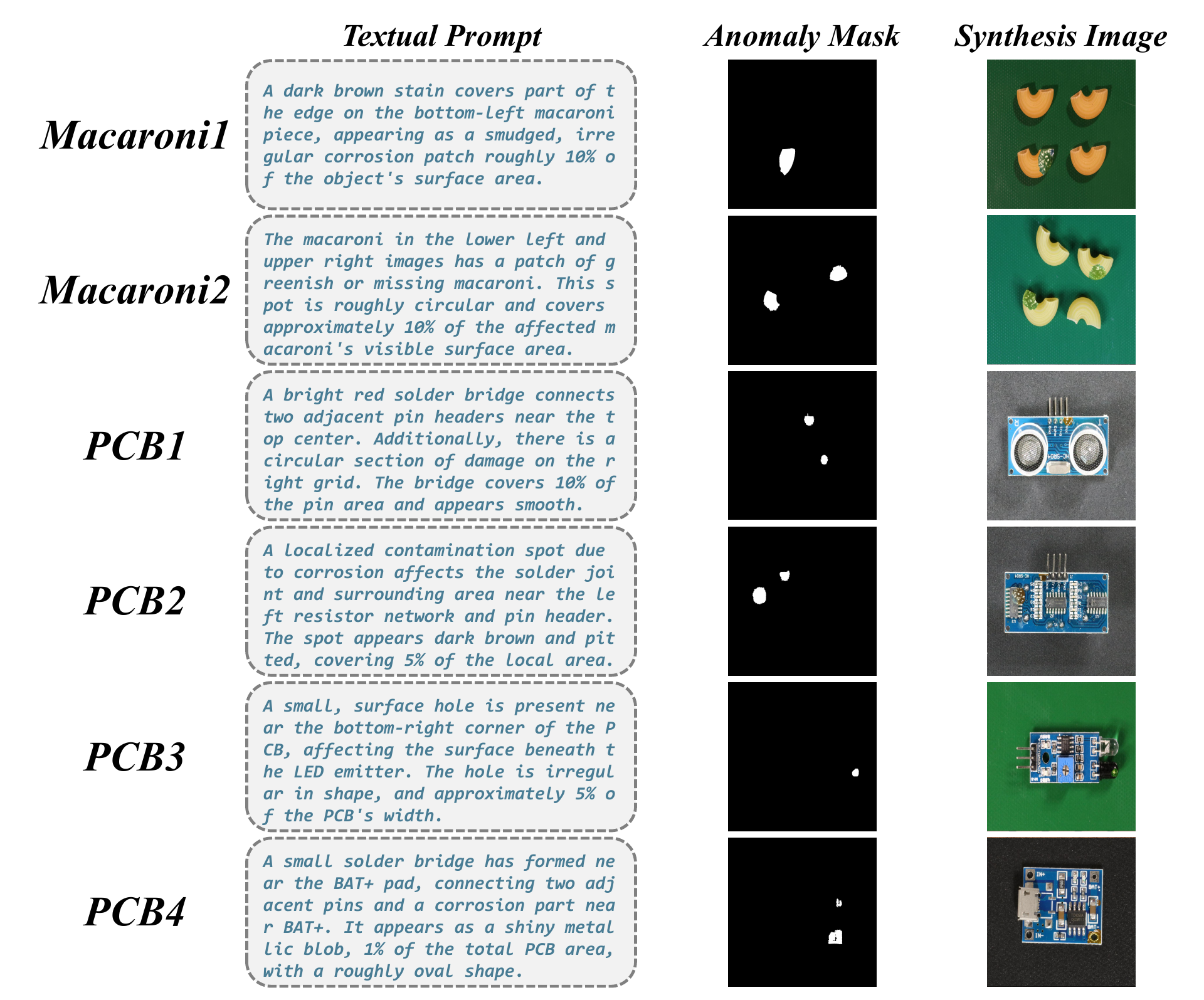}
    \caption{Qualitative results of \textit{ARAS}-generated anomalies and the corresponding textual prompts and anomaly masks on the \textit{VisA} dataset. (Part \textit{II})}
    \label{fig:visa_sys_2}
\end{figure*}

Similar qualitative outcomes are illustrated for the more challenging and recently proposed dataset, \textit{VisA}, in Fig.~\ref{fig:visa_sys_1} and \ref{fig:visa_sys_2}. The generated anomalies span a wide range of appearances, showcasing \textit{ARAS}'s flexibility in capturing complex textures and structural deformations.

\begin{figure*}
    \centering
    \includegraphics[width=1\linewidth]{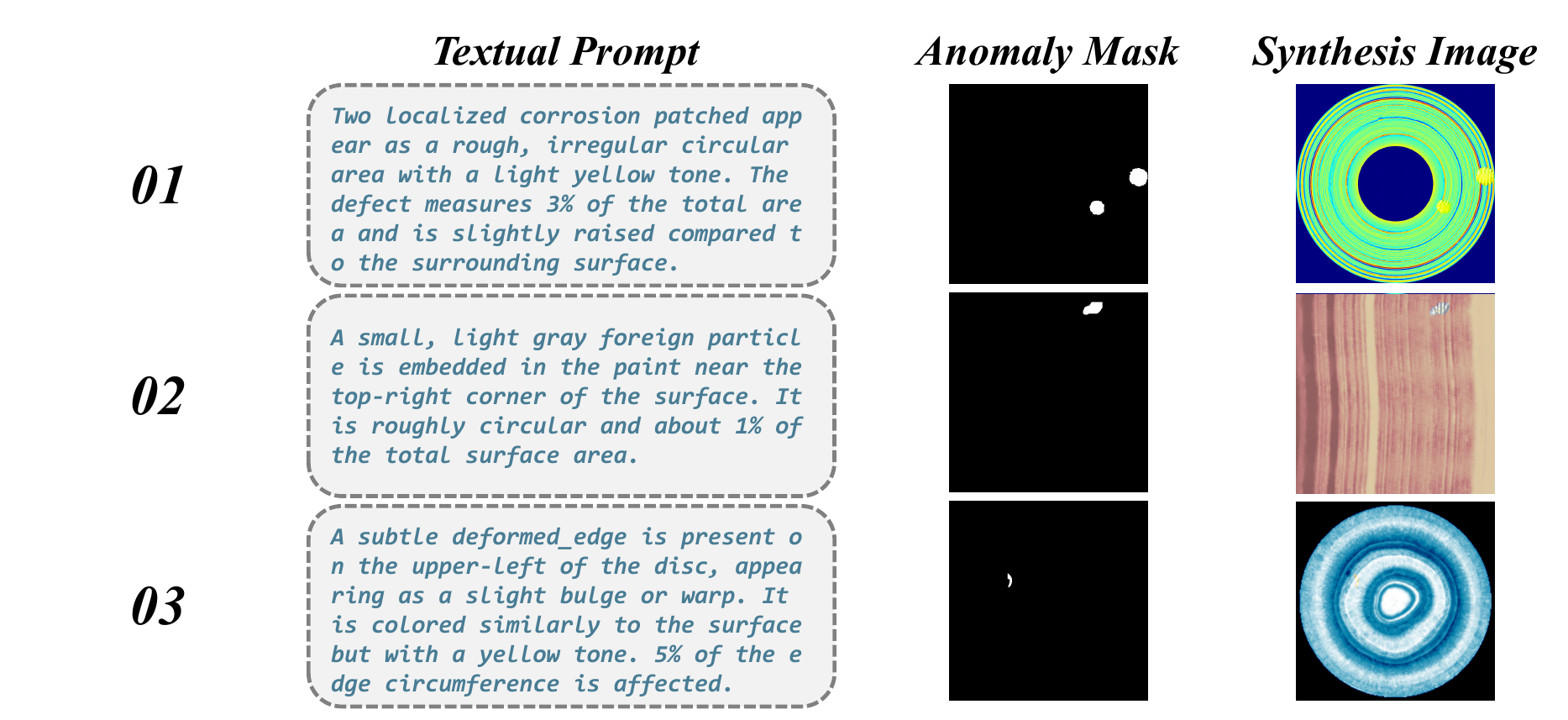}
    \caption{Qualitative results of \textit{ARAS}-generated anomalies and the corresponding textual prompts and anomaly masks on the \textit{BTAD} dataset.}
    \label{fig:btad_sys}
\end{figure*}

In addition, Fig.~\ref{fig:btad_sys} demonstrates anomaly generation performance on the texture-focused dataset \textit{BTAD}, further validating the versatility of \textit{ARAS} across varying industrial contexts.

\subsection{Qualitative Results of \textit{QARAD} Detection Performance}
\begin{figure*}
    \centering
    \includegraphics[width=1\linewidth]{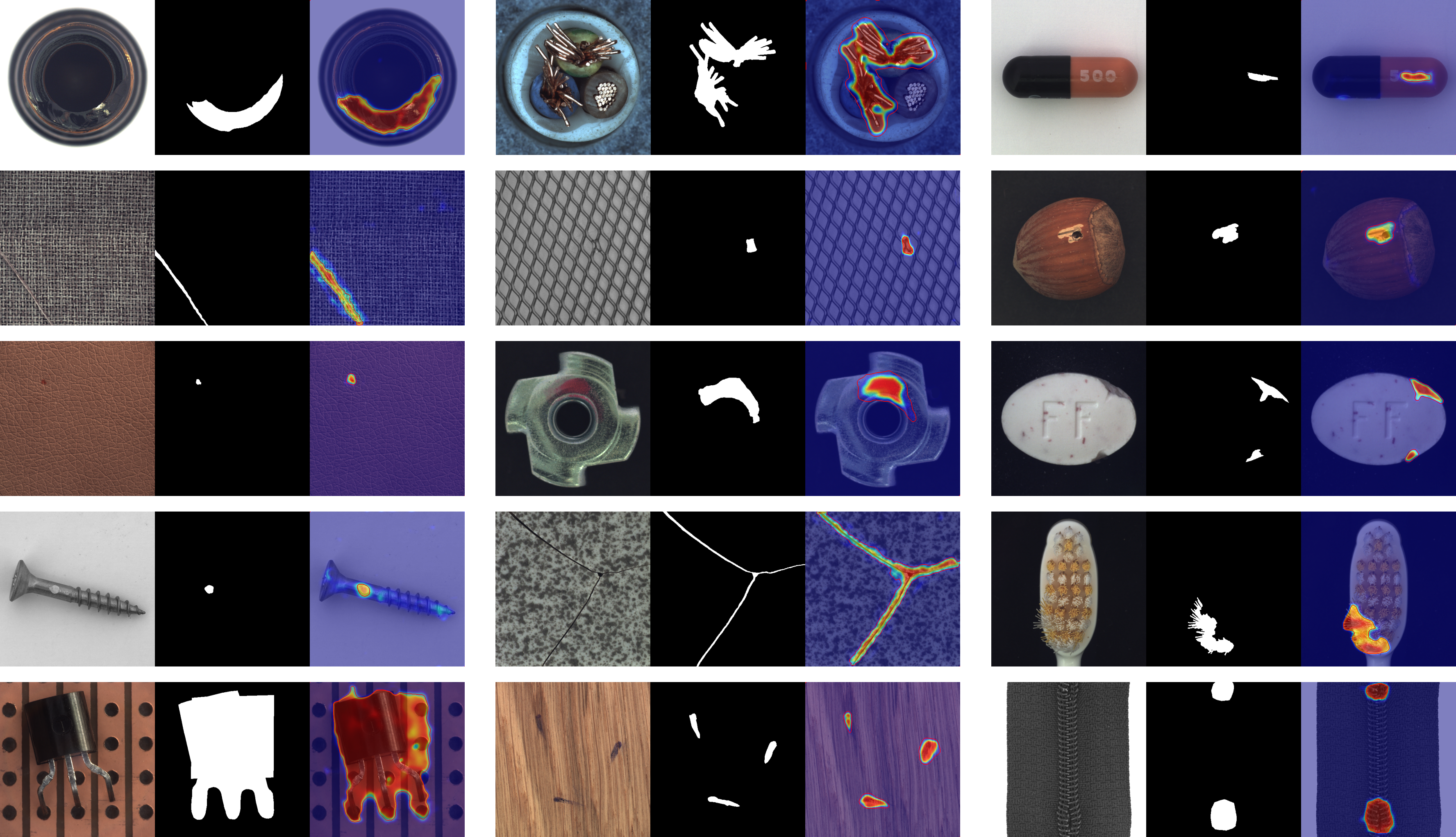}
    \caption{Qualitative results of our anomaly detection method \textit{QARAD} on all \textit{MVTec~AD} categories. Each group shows: (1) original input, (2) ground-truth mask, (3) overlay of detection results.}
    \label{fig:mvtec_result}
\end{figure*}

\begin{figure*}
    \centering
    \includegraphics[width=1\linewidth]{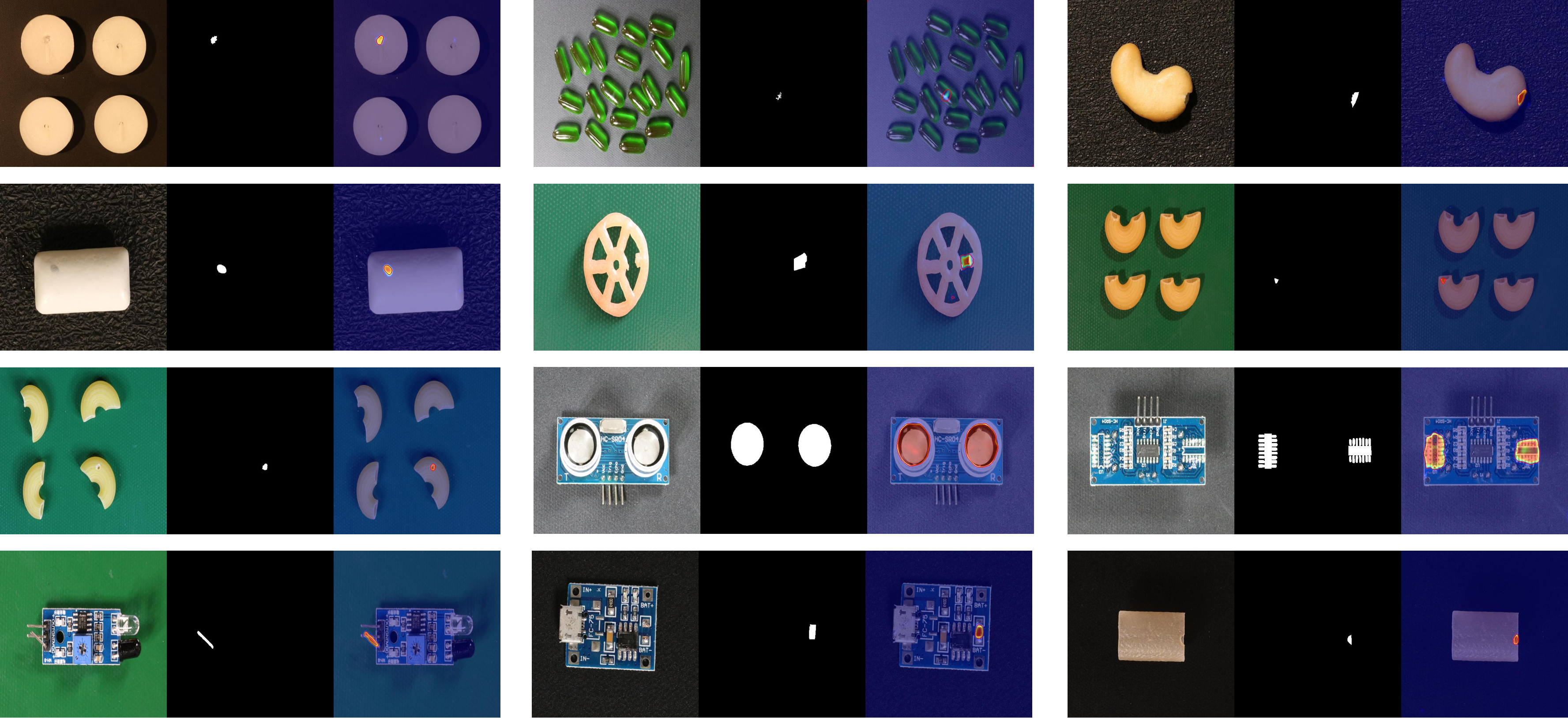}
    \caption{Qualitative results of our anomaly detection method \textit{QARAD} on all \textit{VisA} categories. Each group shows: (1) original input, (2) ground-truth mask, (3) overlay of detection results.}
    \label{fig:visa_result}
\end{figure*}

\begin{figure*}
    \centering
    \includegraphics[width=1\linewidth]{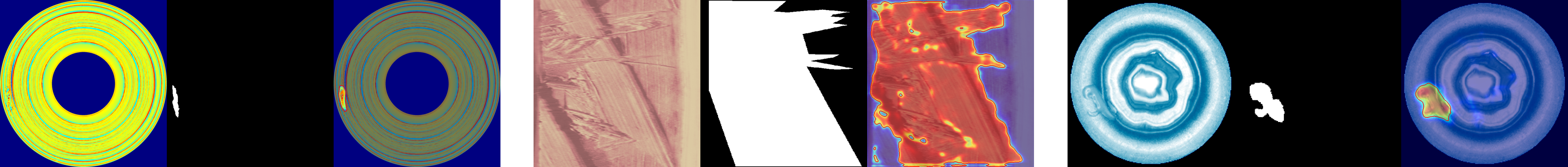}
    \caption{Qualitative results of our anomaly detection method \textit{QARAD} on all \textit{BTAD} categories. Each group shows: (1) original input, (2) ground-truth mask, (3) overlay of detection results.}
    \label{fig:btad_result}
\end{figure*}

Fig.~\ref{fig:mvtec_result}, \ref{fig:visa_result}, and \ref{fig:btad_result} present detection results obtained by our method \textit{QARAD} on the \textit{MVTec~AD}, \textit{VisA}, and \textit{BTAD} datasets, respectively. Each figure shows the original inputs, ground-truth masks, and overlays of our predicted anomaly maps. Notably, \textit{QARAD} accurately identifies subtle anomalies across multiple object categories, confirming its practical suitability for real-world applications.

Overall, these qualitative analyses provide a comprehensive understanding of the anomaly synthesis quality and the detection precision achieved by our proposed pipeline.


\end{document}